\documentclass[lettersize,journal]{IEEEtran}

\usepackage{caption}
\usepackage{amsmath,amsfonts,amsthm,amssymb}
\usepackage{algorithmic}
\usepackage{algorithm}
\usepackage{array}
\usepackage{textcomp}
\usepackage{stfloats}
\usepackage{url}
\usepackage{verbatim}
\usepackage{graphicx}
\usepackage{cite}
\usepackage{booktabs}
\usepackage{multirow}
\usepackage{colortbl}
\usepackage{adjustbox}
\usepackage{pifont}
\usepackage{stfloats}
\usepackage{subfigure}
\usepackage[backref]{hyperref}
\usepackage{threeparttable}
\usepackage{tabularx} 

\begin{document}

\title{Progressive Reliable Supervision for\\Black-Box Universal Domain Adaptation}

\author{
Jing Li, 
Zhe Zhang, 
Huateng Sun, 
Yuhui Yang, 
Wanli Xue,
Xu Cheng,~
Fan Shi,
Jianhua Zhang,
\\Qinghua Hu,~\IEEEmembership{Senior Member,~IEEE},
and Shengyong Chen,~\IEEEmembership{Senior Member,~IEEE}

\thanks{
J. Li, Z. Zhang, H. Sun, Y. Yang, W. Xue, X. Cheng, F. Shi, J. Zhang, and S. Chen are with the School of Computer Science and Engineering, Tianjin University of Technology, Tianjin 300384, China (jing\underline{ }li@tju.edu.cn;
zzhe0124@stud.tjut.edu.cn; 
xuewanli@email.tjut.edu.cn;
xu.cheng@ieee.org;
zjh@ieee.org;
sy@ieee.org
).
Qinghua Hu is with the School of Artificial Intelligence, Tianjin University, Tianjin 300350, China (huqinghua@tju.edu.cn).}
}
\maketitle

\begin{abstract}
Black-Box Universal Domain Adaptation (B$^2$-UniDA) aims to adapt a target model to an unlabeled target domain under arbitrary label-space relationships, while neither source data nor source-model parameters are accessible during adaptation. Under such restrictive information access, reliable target supervision is unavailable, while predictions from the black-box source model can be unreliable due to source bias and domain shifts. To address this challenge, we propose Progressive Reliable Supervision (PReS), which progressively establishes and refines reliable target supervision by coordinating complementary knowledge from the black-box source model, foundation models, and the evolving target model. PReS first identifies reliable known samples by exploiting semantic consistency under diffusion-based perturbations and cross-model consensus between the black-box source model and CLIP. It then estimates the sample-wise reliability of the two predictors and adaptively integrates their complementary predictions to construct reliable pseudo labels. The resulting supervision is exploited asymmetrically to encourage confident and discriminative predictions for reliable known samples while maintaining uncertain yet locally consistent predictions for potential unknown samples. As adaptation proceeds, the emerging target-domain knowledge is further leveraged to refine both the pseudo labels and the known/unknown sample partition, forming a closed-loop process in which supervision construction and target adaptation mutually reinforce each other. Extensive experiments on multiple benchmark datasets under diverse label-space relationships demonstrate that PReS achieves state-of-the-art or highly competitive performance under the restrictive B$^2$-UniDA setting.
\end{abstract}

\begin{IEEEkeywords}
Black-box universal domain adaptation, progressive reliable supervision, foundation models, vision-language models, diffusion models, image classification.
\end{IEEEkeywords}

\section{Introduction}

\IEEEPARstart{U}{nsupervised} domain adaptation (UDA)~\cite{zhang2022transfer,li2025domain,ben2010theory} aims to transfer knowledge from a labeled source domain to an unlabeled target domain under distribution shifts. Conventional UDA typically assumes an identical label space between the two domains, i.e., $\mathcal{Y}^s=\mathcal{Y}^t$, corresponding to closed-set domain adaptation (CDA)~\cite{ganin2015unsupervised,long2015learning,cui2020gradually}. 
In real-world applications, however, diverse category shifts give rise to partial domain adaptation (PDA)~\cite{zhang2018importance,cao2018partial,cao2019learning,WSPDA2024TIP}, open-set domain adaptation (OSDA)~\cite{saito2018open,jing2023WDAN,zeng2025kabon}, and open-partial domain adaptation (OPDA)~\cite{panareda2017open,li2024hyunida}. Universal domain adaptation (UniDA)~\cite{you2019universal} unifies these settings by only assuming the existence of shared classes between the source and target domains. Nevertheless, UniDA requires access to labeled source data during adaptation, which may be infeasible due to privacy, security, and transmission constraints.

Source-free universal domain adaptation (SF-UniDA)~\cite{kundu2020universal,qu2024lead,qu2025glc++} relaxes this requirement by performing adaptation with a pre-trained source model without accessing source data. However, exposing the source model may still raise privacy and intellectual-property concerns. Black-box universal domain adaptation (B$^2$-UniDA)~\cite{deng2021universal} further restricts source-side information by allowing access to neither source data nor source-model parameters; only prediction outputs can be obtained by querying the black-box source model. Together with unknown label-space relationships, these restrictions lead to a fundamental challenge: \emph{reliable target supervision is unavailable during adaptation}.

Existing B$^2$-UniDA methods primarily derive target supervision from black-box source-model predictions~\cite{deng2021universal,he2025gradient,lai2025adu}. However, these predictions suffer from source bias under domain shifts and are particularly unreliable for target-private samples absent from the source label space. Directly using them for pseudo-labeling or known/unknown separation can therefore propagate erroneous supervision. Although existing methods employ confidence-based sample selection, target label-space estimation, or specialized adaptation criteria, effective B$^2$-UniDA still requires constructing trustworthy supervision from limited and imperfect information and progressively improving it during adaptation.

Foundation models~\cite{bommasani2021opportunities} offer complementary knowledge for addressing this challenge. Vision-language models such as CLIP~\cite{radford2021learning} provide generalizable semantic knowledge, while diffusion models~\cite{sohl2015deep,rombach2022high} provide generative priors for producing semantically consistent visual variations. This yields a useful complementarity: the black-box source model provides task-specific but source-biased knowledge, foundation models provide generalizable but task-agnostic priors, and the target model progressively acquires target-domain knowledge during adaptation. Coordinating these knowledge sources therefore provides a promising way to construct and progressively refine reliable target supervision.

Motivated by this insight, we propose \emph{Progressive Reliable Supervision (PReS)}, which progressively constructs and refines reliable target supervision by coordinating complementary knowledge throughout adaptation. PReS first identifies reliable known samples by accumulating semantic-consistency and cross-model-consensus evidence, and then constructs their pseudo labels by adaptively integrating the black-box source-model and CLIP predictions according to sample-wise reliability. The resulting supervision enables asymmetric learning that encourages confident and discriminative predictions for reliable known samples while maintaining uncertain yet locally consistent predictions for potential unknown samples. As the target model acquires target-domain knowledge, its predictions are used to refine the pseudo labels and the known/unknown partition, forming a closed-loop adaptation process.

Our main contributions are summarized as follows:
\begin{itemize}
    \item
    We propose PReS, a progressive reliable-supervision framework for B$^2$-UniDA that coordinates complementary knowledge from the black-box source model, foundation models, and the evolving target model to construct and progressively refine reliable target supervision.

    \item
    We develop a reliable supervision construction mechanism that identifies reliable known samples through diffusion-based semantic consistency and cross-model consensus, and adaptively constructs their pseudo labels according to the sample-wise reliability of the black-box source-model and CLIP predictions.

    \item
    We introduce asymmetric target learning and progressive supervision refinement to effectively exploit and continuously improve the constructed supervision. The emerging target-domain knowledge is fed back to refine both pseudo labels and the known/unknown partition, forming a mutually reinforcing adaptation process.

    \item
    Extensive experiments on multiple benchmark datasets under diverse label-space relationships demonstrate that PReS achieves state-of-the-art or competitive performance under the restrictive B$^2$-UniDA setting.
\end{itemize}

The rest of this paper is organized as follows. Section II reviews related work. Section III presents the problem formulation and PReS. Section IV reports the experiments, and Section V concludes the paper.

\section{Related Work}

\subsection{Universal Domain Adaptation}
Universal domain adaptation (UniDA) addresses domain adaptation without assuming a specific relationship between the source and target label spaces. UAN~\cite{you2019universal} estimates sample transferability through instance-level weighting, while subsequent studies focus on known/unknown separation and target representation learning. DANCE~\cite{saito2020universal} exploits neighborhood clustering to promote discriminative target representations, and OVANet~\cite{saito2021ovanet} performs unknown detection through one-vs-all classifiers. More recently, CCCOP~\cite{zhou2025cross} incorporates cross-domain category context optimization and learnable textual prompts for category-shift adaptation. Despite their effectiveness, these methods require access to labeled source data during adaptation.

\subsection{Source-Free Universal Domain Adaptation}
Recent SFDA studies have explored various aspects of target adaptation, including high-order target structural relations through hypergraph learning~\cite{HGSFDA2025TIP} and adversarial robustness under source-free constraints~\cite{RobustSFDA2026Tip}.
Source-free universal domain adaptation (SF-UniDA)~\cite{kundu2020universal,liang2021umad,wang2025double} performs adaptation using a pre-trained source model and unlabeled target data without accessing source data. UMAD~\cite{liang2021umad} addresses source-free adaptation under universal category shifts, while LEAD~\cite{qu2024lead} exploits singular-value decomposition to characterize transferable and domain-specific components for known/unknown separation. GLC++~\cite{qu2025glc++} exploits target neighborhood structure through contrastive affinity learning, and EGRS~\cite{lu2026exploring} estimates embedding shifts through perturbations and progressively refines target prototypes. Nevertheless, these methods still require access to the parameters and internal representations of the source model, whereas B$^2$-UniDA exposes only its prediction interface.

\subsection{Black-Box Universal Domain Adaptation}
Black-box universal domain adaptation (B$^2$-UniDA)~\cite{deng2021universal} allows access to neither source data nor source-model parameters during adaptation. UB2DA~\cite{deng2021universal} decomposes adaptation into intra-class discrimination and out-of-class detection and identifies potential unknown samples according to prediction uncertainty. ADU~\cite{lai2025adu} modulates supervision according to black-box prediction confidence, although it mainly considers scenarios involving target-private classes. GSS~\cite{he2025gradient} develops a sample-selection strategy for known/unknown separation without a manually predefined threshold. Since these methods primarily derive task-specific supervision from black-box predictions, reliable target supervision remains difficult to establish when such predictions are corrupted by domain shifts or target-private samples.

Different from these approaches, PReS coordinates complementary knowledge from the black-box source model, foundation models, and the evolving target model to progressively construct and refine reliable supervision throughout adaptation.

\subsection{Foundation Models for Domain Adaptation}
Foundation models~\cite{bommasani2021opportunities} pretrained on large-scale data provide rich prior knowledge for downstream visual tasks. Vision-language models such as CLIP~\cite{radford2021learning} learn transferable visual-semantic representations from large-scale image-text data, while subsequent models such as BLIP~\cite{li2022blip}, Flamingo~\cite{alayrac2022flamingo}, and FLAVA~\cite{singh2022flava} further demonstrate the generalization capabilities of multimodal pretraining. Recent studies have further exploited such vision-language priors for domain adaptation, e.g., by coordinating textual and visual prompt tuning for domain alignment \cite{CuiUDA2025Tip} or progressively integrating teacher and student predictions to refine target pseudo-labels \cite{PHPLUDA2026Tip}.
Such semantic priors complement the task-specific but potentially source-biased predictions of a black-box source model.

Diffusion models~\cite{sohl2015deep} provide another form of external prior through their strong generative capabilities. Stable Diffusion~\cite{rombach2022high}, in particular, can generate diverse visual variations while preserving high-level semantic content. PReS exploits these complementary priors for reliable supervision construction: CLIP provides semantic knowledge for cross-model consensus and reliability-aware pseudo-label generation, while Stable Diffusion provides semantic-preserving perturbations for evaluating prediction consistency.

\section{Method}

\subsection{Problem Definition}

\begin{figure*}[!t]
\centering
\includegraphics[width=1\textwidth]{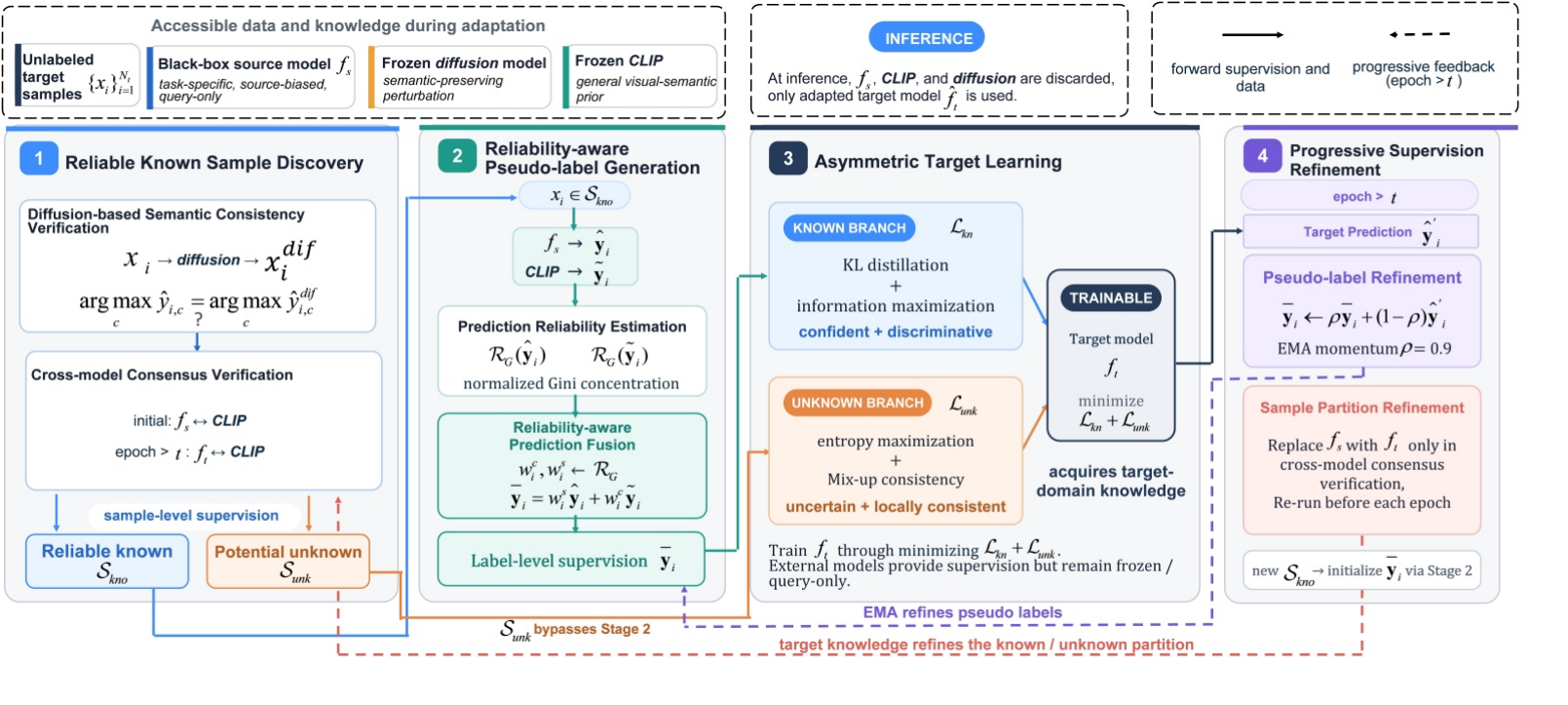}
\caption{Overview of the proposed PReS framework for B$^{2}$-UniDA. PReS progressively constructs sample- and label-level supervision from complementary knowledge sources, performs asymmetric target learning, and exploits the emerging target-domain knowledge to refine the supervision throughout adaptation. Solid and dashed arrows denote the forward adaptation flow and progressive feedback, respectively. Only the adapted target model is required during inference.}
\label{overview}
\end{figure*}

B$^{2}$-UniDA considers a highly restricted adaptation setting where only an unlabeled target domain
$D_t=\{x_i\}_{i=1}^{N_t}$, $x_i\in\mathcal{X}^t$, and prediction outputs obtained by querying a black-box source model $f_s$ are available. Neither the source data nor the architecture and parameters of $f_s$ are accessible.
Let $\mathcal{Y}^s$ and $\mathcal{Y}^t$ denote the source and target label spaces, respectively, with $|\mathcal{Y}^s|=C$. We define the shared, source-private, and target-private label spaces as
\begin{equation}
\mathcal{Y}=\mathcal{Y}^s\cap\mathcal{Y}^t,\quad
\overline{\mathcal{Y}}^s=\mathcal{Y}^s\setminus\mathcal{Y}^t,\quad
\overline{\mathcal{Y}}^t=\mathcal{Y}^t\setminus\mathcal{Y}^s.
\end{equation}
Following UniDA, we assume $\mathcal{Y}^s\cap\mathcal{Y}^t\neq\varnothing$, without imposing any additional relationship between $\mathcal{Y}^s$ and $\mathcal{Y}^t$.

The objective is to learn a target model $f_t$ using only the target samples and black-box predictions, such that samples from the shared label space $\mathcal{Y}$ are correctly classified, while those from the target-private label space $\overline{\mathcal{Y}}^t$ are recognized as a single unknown class.

\subsection{Method Overview}
The overall framework of our proposed Progressive Reliable Supervision (PReS) is illustrated in Fig.~\ref{overview}.
Under the B$^{2}$-UniDA setting, reliable target supervision is unavailable, while the accessible knowledge sources exhibit complementary strengths and limitations. The black-box source model provides task-specific but source-biased knowledge, foundation models offer generalizable but task-agnostic priors, and the target model can progressively acquire target-domain knowledge but is unreliable at the beginning of adaptation. 
PReS coordinates these complementary knowledge sources to progressively establish reliable supervision during adaptation.

Our framework consists of four stages. First, \emph{Reliable Known Sample Discovery} identifies reliable known samples through diffusion-based semantic consistency and cross-model consensus. Second, \emph{Reliability-aware Pseudo-label Generation} adaptively integrates the black-box source-model and CLIP predictions according to their sample-wise reliability. Third, \emph{Asymmetric Target Learning} exploits the resulting supervision according to the distinct roles of reliable known and potential unknown samples. Finally, \emph{Progressive Supervision Refinement} leverages the emerging target-domain knowledge to refine both pseudo labels and the known/unknown sample partition.

These stages form a closed-loop adaptation process: complementary external knowledge first establishes reliable supervision to facilitate target adaptation, while the emerging target knowledge is subsequently fed back to refine the supervision itself. Consequently, supervision construction and target adaptation mutually reinforce each other throughout training.

\subsection{Reliable Known Sample Discovery (Stage 1)}
\label{RKSD}
Under the B$^{2}$-UniDA setting, predictions from the black-box source model are inevitably affected by domain shift and therefore cannot be directly adopted as reliable supervision. We identify reliable known samples by progressively accumulating two complementary sources of \emph{reliability evidence}. Specifically, reliable known samples are expected to exhibit stable predictions under semantic-preserving perturbations and consistent semantic understanding across complementary vision models. Accordingly, we first construct \emph{semantic consistency evidence} through Diffusion-based Semantic Consistency Verification and then further validate candidate known samples using \emph{cross-model consensus evidence} through Cross-model Consensus Verification (CCV).

\subsubsection{Diffusion-based Semantic Consistency Verification}
We first evaluate whether the source-model prediction of a target sample remains stable under semantic-preserving perturbations. Unlike conventional image augmentations that mainly introduce predefined low-level appearance variations, diffusion models can generate visually diverse yet semantically consistent images, making them more suitable for evaluating semantic prediction consistency.

Specifically, each target sample $x_i$ is fed into a diffusion pipeline together with a fixed text prompt~\cite{rombach2022high}. The positive prompt is defined as ``preserve original content and structure, sharp focus, fine texture details, natural lighting'', while the negative prompt suppresses ``blurriness, noise, artifacts, changes in content or composition, and stylized distortions''. Under these constraints, the diffusion model generates a semantic-preserving counterpart $x_i^{dif}$.

The black-box source model $f_s$ is then queried to predict both $x_i$ and $x_i^{dif}$, yielding the corresponding predictive probability vectors $\hat{\mathbf{y}}_i \in \mathbb{R}^C$ and $\hat{\mathbf{y}}_i^{dif} \in \mathbb{R}^C$, respectively.
When $\mathop{\arg\max}_{c}\hat{y}_{i,c}=\mathop{\arg\max}_{c}\hat{y}_{i,c}^{dif}$, the sample is retained as a candidate known sample; otherwise, it is regarded as a potential unknown sample. Since the resulting candidate known set may still contain false positives owing to source-domain bias, we further verify these samples using complementary semantic knowledge from CLIP.

\subsubsection{Cross-model Consensus Verification}
For each candidate known sample, CLIP produces the predictive probability vector $\tilde{\mathbf{y}}_i \in \mathbb{R}^C$. The sample is accepted as a reliable known sample only when the predictions of the black-box source model and CLIP agree, i.e.,
$\mathop{\arg\max}_{c} \hat{y}_{i,c} = \mathop{\arg\max}_{c} \tilde{y}_{i,c}.$
Otherwise, the sample is regarded as a potential unknown sample and is assigned to the unknown-sample set.

After the two successive verification stages, the reliable known-sample set and unknown-sample set are respectively defined as
\begin{equation}
\label{xkn}
\mathcal{S}_{kno}=\{x_i^{kno}\}_{i=1}^{N_{kno}},
\end{equation}
and
\begin{equation}
\label{xunk}
\mathcal{S}_{unk}=\{x_i^{unk}\}_{i=1}^{N_{unk}},
\end{equation}
where $N_{kno}$ and $N_{unk}$ denote the numbers of reliable known samples and potential unknown samples, respectively.

\subsection{Reliability-aware Pseudo-label Generation (Stage 2)}
Although $\mathcal{S}_{kno}$ contains reliable known samples, their predicted classes may still be incorrect due to source-domain bias of the black-box source model and the limited task-specific discrimination of CLIP. Since the two predictors provide complementary task-specific and general semantic knowledge, we estimate their sample-wise reliability and adaptively fuse their predictions to construct reliable label-level supervision.

\subsubsection{Prediction Reliability Estimation}
For each known sample $x_i^{kno}\in\mathcal{S}_{kno}$, $\hat{\mathbf{y}}_i$ and $\tilde{\mathbf{y}}_i$ denote the predictive probability vectors produced by the black-box source model and CLIP, respectively. To uniformly characterize their prediction reliability, we consider a generic predictive probability vector
\begin{equation}
\mathbf{q}_i=[q_i^1,\ldots,q_i^C],
\qquad
q_i^c\geq0,\quad
\sum_{c=1}^{C}q_i^c=1.
\end{equation}

A reliable prediction typically concentrates its probability mass on a small number of classes, whereas an ambiguous prediction is more uniformly distributed. We therefore employ the Gini coefficient~\cite{stark2024gini,eliazar2024beautiful} to quantify probability concentration as a sample-wise prediction reliability measure.

Let $q_i^{(1)}\leq q_i^{(2)}\leq\cdots\leq q_i^{(C)}$ denote the elements of $\mathbf{q}_i$ sorted in ascending order. We define the normalized Gini-based prediction reliability as
\begin{equation}
\label{eq:gini_reliability}
\mathcal{R}_{G}(\mathbf{q}_i) = 
\frac{C}{C-1}
\left(
1-
2\sum_{c=1}^{C}
q_i^{(c)}
\frac{C-c+\frac{1}{2}}{C}
\right).
\end{equation}
Since the standard Gini coefficient of a $C$-dimensional probability vector ranges from $0$ to $(C-1)/C$, the factor $C/(C-1)$ normalizes $\mathcal{R}_{G}(\mathbf{q}_i)$ to $[0,1]$, eliminating the dependence of its maximum value on the number of classes. A uniform distribution yields $\mathcal{R}_{G}(\mathbf{q}_i)=0$, indicating maximum predictive ambiguity, whereas a one-hot distribution yields $\mathcal{R}_{G}(\mathbf{q}_i)=1$, indicating maximum probability concentration. Accordingly, $\mathcal{R}_{G}(\hat{\mathbf{y}}_i)$ and $\mathcal{R}_{G}(\tilde{\mathbf{y}}_i)$ quantify the sample-wise reliability of the black-box source model and CLIP, respectively.

\subsubsection{Reliability-aware Prediction Fusion}
Rather than using fixed fusion weights, we determine the contribution of each predictor according to its relative reliability for each target sample. Specifically, the reliability-aware weight assigned to the black-box source model is defined as
\begin{equation}
\label{eq:fusion_weight}
w_i^{s} = 
\frac{
\mathcal{R}_{G}(\hat{\mathbf{y}}_i) + \epsilon}
{\mathcal{R}_{G}(\hat{\mathbf{y}}_i) + \mathcal{R}_{G}(\tilde{\mathbf{y}}_i) + 2\epsilon},
\end{equation}
where $\epsilon$ is a small positive constant introduced for numerical stability. The corresponding weight for CLIP is defined as
$w_i^{c}=1-w_i^{s}$.
The reliability-aware fused prediction is then obtained as
\begin{equation}
\label{eq:reliable_pseudo}
\bar{\mathbf{y}}_i = w_i^{s}\hat{\mathbf{y}}_i + w_i^{c}\tilde{\mathbf{y}}_i.
\end{equation}

Thus, the predictor with higher sample-wise reliability contributes more strongly to the fused prediction, adaptively exploiting the complementary knowledge of the black-box source model and CLIP. For each $x_i^{kno}\in\mathcal{S}_{kno}$, $\bar{\mathbf{y}}_i$ provides label-level supervision for subsequent target adaptation.

\subsection{Asymmetric Target Learning (Stage 3)}
The preceding stages partition target samples into reliable known samples $\mathcal{S}_{kno}$ and potential unknown samples $\mathcal{S}_{unk}$, which require distinct learning objectives. Reliable known samples should provide discriminative supervision for learning shared classes, whereas potential unknown samples should not be forced toward any source class. Accordingly, we encourage reliable known samples to be \emph{confident and discriminative}, while maintaining \emph{uncertain yet locally consistent} predictions for potential unknown samples.

\subsubsection{Known-sample Knowledge Transfer}
For each known sample $x_i^{kno}\in\mathcal{S}_{kno}$, let $\hat{\mathbf{y}}'_i=f_t(x_i^{kno})$ denote the predictive probability vector produced by the target model. To transfer the reliable supervision encoded in the fused prediction $\bar{\mathbf{y}}_i$, we employ the knowledge distillation loss~\cite{hinton2015distilling}
\begin{equation}
\label{losskd}
\mathcal{L}_{kd}=
\mathbb{E}_{x_i^{kno}\in\mathcal{S}_{kno}}
D_{KL}
\left(
\bar{\mathbf{y}}_i
\parallel
\hat{\mathbf{y}}'_i
\right),
\end{equation}
where $D_{KL}(\cdot\parallel\cdot)$ denotes the KL divergence.

To encourage confident predictions while preserving class diversity, we further introduce information maximization regularization~\cite{liang2020we,krause2010discriminative}:
\begin{equation}
\label{lossim}
\mathcal{L}_{im}=
H\left(
\mathbb{E}_{x_i^{kno}\in\mathcal{S}_{kno}}
\hat{\mathbf{y}}'_i
\right)
-
\mathbb{E}_{x_i^{kno}\in\mathcal{S}_{kno}}
H\left(
\hat{\mathbf{y}}'_i
\right),
\end{equation}
where $H(\cdot)$ denotes the Shannon entropy. Maximizing the first term promotes diversity in the overall predictive class distribution, while minimizing the second encourages confident predictions for individual samples. The overall objective for reliable known samples is
\begin{equation}
\label{losskn}
\mathcal{L}_{kn}=\mathcal{L}_{kd} - \mathcal{L}_{im},
\end{equation}
which is minimized during training.

\subsubsection{Unknown-sample Uncertainty Regularization}
For potential unknown samples $x_i^{unk}\in\mathcal{S}_{unk}$, no reliable class-specific supervision is available. We therefore encourage high predictive uncertainty over the source classes through the negative entropy loss
\begin{equation}
\label{lossem}
\mathcal{L}_{em} = 
\frac{1}{N_{unk}}
\sum_{i=1}^{N_{unk}}
\sum_{c=1}^{C}
\hat{y}'_{i,c}
\log \hat{y}'_{i,c},
\end{equation}
where $\hat{y}'_{i,c}$ denotes the probability predicted by $f_t$ that the $i$-th unknown sample belongs to the $c$-th source class. Minimizing $\mathcal{L}_{em}$ maximizes predictive entropy, discouraging confident assignment of potential unknown samples to source classes.

Since entropy maximization does not explicitly enforce local prediction smoothness, we further introduce Mixup-based consistency regularization. Given $x_i^{unk},x_j^{unk}\in\mathcal{S}_{unk}$, we define
\begin{equation}
\mathrm{Mix}_{\lambda}(a,b) = 
\lambda a+(1-\lambda)b,
\end{equation}
where $\lambda\sim\mathrm{Beta}(0.3,0.3)$. The Mixup consistency loss is formulated as
\begin{align}
\label{lossmix}
\mathcal{L}_{{mix}}
&= {\mathbb{E}}_{x_i^{{unk}}, x_j^{{unk}} \in \mathcal{S}_{{unk}}}
{\mathbb{E}}_{\lambda \sim \mathrm{Beta}(0.3, 0.3)} \nonumber \\
&\quad D_{{KL}}\Big(
\mathrm{Mix}_{\lambda} \big(\hat{\mathbf{y}}_i', \hat{\mathbf{y}}_j'\big)
\;\big\|\;
f_t\big(\mathrm{Mix}_{\lambda}(x_i^{{unk}}, x_j^{{unk}})\big)
\Big).
\end{align}
This objective encourages locally consistent predictions for potential unknown samples under input interpolation.

The overall objective for potential unknown samples is
\begin{equation}
\label{lossunk}
\mathcal{L}_{unk}=
\beta\mathcal{L}_{em}
+
\gamma\mathcal{L}_{mix},
\end{equation}
where $\beta$ and $\gamma$ control the contributions of entropy maximization and consistency regularization, respectively.

\subsection{Progressive Supervision Refinement (PSR, Stage 4)} \label{update}
The supervision constructed in the preceding stages is initially derived from the black-box source model and foundation models and therefore remains imperfect under domain shift. As adaptation proceeds, the target model progressively acquires target-domain knowledge, making its predictions increasingly informative for refining the supervision. Accordingly, after an initial adaptation period of $t$ epochs, we exploit the emerging target knowledge to progressively refine both the pseudo labels and the known/unknown sample partition.

\subsubsection{Pseudo-label Refinement (PLR)}
To progressively incorporate target-domain knowledge into label-level supervision, we update the pseudo label $\bar{\mathbf{y}}_i$ using the target prediction $\hat{\mathbf{y}}'_i$ through EMA. Specifically, at the end of each epoch after the initial $t$ training epochs, the pseudo label of each known sample is updated as
\begin{equation}
\label{mempupdate}
\bar{\mathbf{y}}_i
\leftarrow
\rho\,\bar{\mathbf{y}}_i
+
(1-\rho)\,\hat{\mathbf{y}}'_i,
\end{equation}
where $\rho$ denotes the EMA momentum coefficient and is empirically set to $0.9$.
This update gradually incorporates target-specific knowledge while retaining the reliable information in the existing pseudo label, with temporal smoothing reducing the influence of potentially noisy instantaneous predictions.

\subsubsection{Sample Partition Refinement (SPR)} \label{Sample Partition Refinement}
To refine the sample-level supervision, starting from the $(t+1)$-th epoch, we replace the black-box source model with the target model \emph{only in Cross-model Consensus Verification}, while keeping Diffusion-based Semantic Consistency Verification unchanged. Specifically, the first verification stage continues to identify candidate known samples according to the consistency between $\hat{\mathbf{y}}_i$ and $\hat{\mathbf{y}}_i^{dif}$ produced by the black-box source model. For each candidate known sample, the target prediction $\hat{\mathbf{y}}'_i$ is then used in place of $\hat{\mathbf{y}}_i$ to evaluate its consensus with the CLIP prediction $\tilde{\mathbf{y}}_i$. Before each subsequent training epoch, the cached semantic-consistency evidence is reused, while Cross-model Consensus Verification is updated using the target-model predictions to refine $\mathcal{S}_{kno}$ and $\mathcal{S}_{unk}$.

As the partition evolves, previously potential unknown samples may be newly admitted into $\mathcal{S}_{kno}$. For each newly discovered known sample, we initialize its pseudo label by re-applying Reliability-aware Pseudo-label Generation, where the black-box source-model and CLIP predictions are fused according to their sample-wise reliability using equation~(\ref{eq:reliable_pseudo}). Samples that remain in $\mathcal{S}_{kno}$ retain their progressively refined pseudo labels from previous epochs. Thus, every sample in the updated known set is equipped with label-level supervision before subsequent target-model optimization.

\subsection{Training and Inference}
The target model is jointly optimized over the reliable known and potential unknown samples as
\begin{equation}
\label{overall_objective}
\hat{f}_t=\mathop{\arg\min}_{f_t}
\left(
\mathcal{L}_{kn}
+
\mathcal{L}_{unk}
\right),
\end{equation}
where $\mathcal{L}_{kn}$ and $\mathcal{L}_{unk}$ are defined in Eqs.~(\ref{losskn}) and~(\ref{lossunk}), respectively. The complete training procedure is summarized in Algorithm~\ref{alg}.

\begin{algorithm}[t!]
\caption{Training Procedure of PReS}
\label{alg}
\renewcommand{\algorithmicrequire}{\textbf{Input:}}
\renewcommand{\algorithmicensure}{\textbf{Output:}}
\begin{algorithmic}[1]
\REQUIRE Unlabeled target domain $D_t=\{x_i\}_{i=1}^{N_t}$; query access to $f_s$; target model $f_t$; CLIP and diffusion models; $\beta$, $\gamma$, $T$, and $t$.
\ENSURE Adapted target model $\hat{f}_t$.

\STATE Perform Stage 1 to obtain $\mathcal{S}_{kno}$ and $\mathcal{S}_{unk}$.
\STATE Perform Stage 2 to initialize $\bar{\mathbf{y}}_i$ for each $x_i^{kno}\in\mathcal{S}_{kno}$.

\FOR{$epoch=1$ to $T$}
\IF{$epoch>t$}
    \STATE Store $\mathcal{S}_{kno}^{old}\leftarrow\mathcal{S}_{kno}$ and perform SPR to update $\mathcal{S}_{kno}$ and $\mathcal{S}_{unk}$.
    \STATE Initialize $\bar{\mathbf{y}}_i$ via Eq.~(\ref{eq:reliable_pseudo}) for each $x_i^{kno}\in\mathcal{S}_{kno}\setminus\mathcal{S}_{kno}^{old}$.
\ENDIF
\STATE Compute $\mathcal{L}_{kn}$ and $\mathcal{L}_{unk}$ via Eqs.~(\ref{losskn}) and~(\ref{lossunk}), respectively.
\STATE Update $f_t$ by minimizing $\mathcal{L}_{kn}+\mathcal{L}_{unk}$.
\IF{$epoch>t$}
    \STATE Perform PLR via Eq.~(\ref{mempupdate}).
\ENDIF
\ENDFOR

\STATE \textbf{return} $\hat{f}_t$.
\end{algorithmic}
\end{algorithm}

During inference, only the adapted target model $\hat{f}_t$ is required. Following GLC~\cite{qu2023upcycling}, we use normalized Shannon entropy to quantify predictive uncertainty:
\begin{equation}
\label{InferenceEntropy}
H(x_i)=
-\frac{1}{\log C}
\sum_{c=1}^{C}
\hat{y}'_{i,c}
\log \hat{y}'_{i,c}.
\end{equation}
The final prediction is determined as
\begin{equation}
y(x_i)=
\begin{cases}
\mathrm{unknown}, & H(x_i)\geq w,\\
\mathop{\mathrm{argmax}}\limits_{c\in\{1,\ldots,C\}} \hat{y}'_{i,c}, & H(x_i)<w,
\end{cases}
\label{eq:final_prediction}
\end{equation}
where $w$ denotes the uncertainty threshold for unknown-class detection.

\section{EXPERIMENTS}

\begin{table}[!t]
\centering
\begin{threeparttable}
\caption{Class splits of each dataset under different domain adaptation scenarios. $\mathcal{Y}$, $\overline{\mathcal{Y}}^s$, and $\overline{\mathcal{Y}}^t$ denote the shared label set between the source and target domains, the source private label set, and the target private label set, respectively.}
\begin{tabularx}{\columnwidth}{l | X X X}   
\toprule
\multirow{2}{*}{Dataset} & \multicolumn{3}{c}{Class Split ($\mathcal{Y}/\overline{\mathcal{Y}}^s/\overline{\mathcal{Y}}^t$)} \\
\cmidrule(lr){2-4}
              & OPDA       & OSDA & PDA  \\
\midrule
Office-31     & 10/10/11   & 10/0/11 & 10/21/0 \\
Office-Home   & 10/5/50    & 25/0/40 & 25/40/0 \\
VisDA         & 6/3/3      & 6/0/6   & 6/6/0   \\
DomainNet     & 150/50/145 & -       & -       \\
\bottomrule
\end{tabularx}
\label{classnum}
\end{threeparttable}
\end{table}

\begin{table*}[htb]
\centering
\caption{H-score (\%) comparison under OPDA on Office-31, VisDA, and DomainNet. Italic, underlined, and bold denote the best UniDA, SF-UniDA, and B$^2$-UniDA results, respectively. ‘–’ denotes unavailable results.}
\begin{adjustbox}{max width=\textwidth}
\begin{tabular}{c | c | c c c c c c c | c | c c c c c c c }
\toprule
\multirow{2}{*}{\textbf{Type}} & \multirow{2}{*}{\textbf{Method}} & \multicolumn{7}{c|}{\textbf{Office-31 (10/10/11)}} & \textbf{VisDA (6/3/3)} & \multicolumn{7}{c}{\textbf{DomainNet (150/50/145)}} \\
\cmidrule(lr){3-9}
\cmidrule(lr){10-10}
\cmidrule(lr){11-17}
& & A$\rightarrow$D & A$\rightarrow$W & D$\rightarrow$A & D$\rightarrow$W & W$\rightarrow$A & W$\rightarrow$D & \textbf{Avg.} & S$\rightarrow$R & P$\rightarrow$R & P$\rightarrow$S & R$\rightarrow$P & R$\rightarrow$S & S$\rightarrow$P & S$\rightarrow$R & \textbf{Avg.} \\
\midrule
\multirow{5}{*}{UniDA}
 & UAN\cite{you2019universal}    & 59.7 & 58.6 & 60.1 & 70.6 & 60.3 & 71.4 & 63.5 & 34.8 & 41.9 & 39.1 & 43.6 & 38.7 & 38.9 & 43.7 & 41.0\\
 & DANCE\cite{saito2020universal}  & 78.6 & 71.5 & \emph{79.9} & 91.4 & 72.2 & 87.9 & 80.3 & 42.8 & 21.0 & 37.0 & 47.3 & 46.7 & 27.7 & 21.0 & 33.5\\
 & DCC\cite{li2021domain}    & \emph{88.5} & 78.5 & 70.2 & 79.3 & 75.9 & 88.6 & 80.2 & 43.0 & 56.9 & 43.7 & 50.3 & 43.3 & 44.9 & 56.2 & 49.2\\
 & OVANet\cite{saito2021ovanet} & 84.1 & \emph{79.3} & 78.5 & \emph{94.3} & \emph{83.5} & \emph{95.8} & \emph{85.9} & 48.1 & 55.4 & 45.8 & 52.1 & 44.8 & 46.3 & 54.5 & 49.8\\
 & CCCOP\cite{zhou2025cross}  & - & - & - & - & - & - & - & \emph{76.7} & \emph{69.2} & \emph{54.2} & \emph{55.5} & \emph{54.5} & \emph{56.1} & \emph{70.2} & \emph{60.0}\\
\midrule
\multirow{5}{*}{SF-UniDA} 
 & UMAD\cite{liang2021umad}     & \underline{86.9} & \underline{88.3} & 90.9 & \underline{95.9} & \underline{90.0} & \underline{98.0} & \underline{91.7} & 55.1 & 53.4 & 39.5 & 44.1 & 37.6 & 40.5 & 52.3 & 44.6\\
 & LEAD\cite{qu2024lead}     & 85.4 & 85.0 & 86.3 & 90.9 & 86.2 & 93.1 & 87.8 & 76.6 & 59.9 & 46.1 & 51.3 & 45.0 & 45.9 & 56.3 & 50.8\\
 & GLC++\cite{qu2025glc++}    & 81.7 & 85.3 & 89.7 & 91.5 & 88.1 & 92.6 & 88.2 & 75.3 & \underline{62.8} & \underline{50.5} & \underline{55.0} & \underline{50.5} & \underline{49.0} & \underline{60.9} & \underline{54.8}\\
 & EGRS\cite{lu2026exploring}     & 82.2 & 87.9 & \underline{91.1} & 92.0 & 89.7 & 92.9 & 89.3 & 76.7 & - & - & - & - & - & - & -\\
 & ESBN\cite{li2025esbn}     & 85.3 & 83.9 & 86.9 & 94.0 & 87.3 & 94.6 & 88.7 & \underline{78.1} & - & - & - & - & - & - & -\\
\midrule
\multirow{5}{*}{B$^2$-UniDA}
 & SO         & 86.3 & 71.0 & 33.0 & 69.1 & 43.4 & 81.9 & 64.1 & 41.2 & \textbf{63.9} & 46.7 & 53.2 & 46.3 & 38.0 & 55.7 & 50.6\\
 & UB2DA\cite{deng2021universal}      & 80.9 & 78.2 & \textbf{92.6} & 92.6 & 89.4 & 87.9 & 86.9 & 45.2 & 57.1 & 47.2 & 54.8 & 44.0 & 41.4 & 51.5 & 49.3\\
 & ADU\cite{lai2025adu}        & 87.5 & 85.2 & 87.0 & \textbf{94.4} & 83.8 & 90.5 & 88.1 & 48.7 & 59.8 & 47.8 & 52.5 & 46.6 & 42.7 & 56.4 & 51.0\\
 & GSS\cite{he2025gradient}        & 81.1 & 80.9 & 91.4 & 92.0 & 88.1 & 89.4 & 87.1 & 43.5 & 59.1 & 46.7 & 52.1 & 45.0 & 43.9 & 55.2 & 50.3\\
 & PReS(ours) & \textbf{89.5} & \textbf{90.7} & 89.7 & 92.8 & \textbf{91.4} & \textbf{93.1} & \textbf{91.2} & \textbf{66.5} & 54.9 & \textbf{56.8} & \textbf{59.5} & \textbf{56.2} & \textbf{56.8} & \textbf{66.2} & \textbf{58.4}\\
\bottomrule
\end{tabular}
\end{adjustbox}
\label{office31op}
\end{table*}

\begin{table*}[htb]
\centering
\caption{H-score (\%) comparison under OPDA on Office-Home. Italic, underlined, and bold denote the best UniDA, SF-UniDA, and B$^2$-UniDA results, respectively.}
\begin{adjustbox}{max width=\textwidth}
\begin{tabular}{c | c | c c c c c c c c c c c c c }
\toprule
\multirow{2}{*}{\textbf{Type}} & \multirow{2}{*}{\textbf{Method}} & \multicolumn{13}{c}{\textbf{Office-Home (10/5/50)}}\\
\cmidrule(lr){3-15}
& & Ar$\rightarrow$Cl & Ar$\rightarrow$Pr & Ar$\rightarrow$Rw & Cl$\rightarrow$Ar & Cl$\rightarrow$Pr & Cl$\rightarrow$Rw & Pr$\rightarrow$Ar & Pr$\rightarrow$Cl &  Pr$\rightarrow$Rw & Rw$\rightarrow$Ar & Rw$\rightarrow$Cl & Rw$\rightarrow$Pr & \textbf{Avg.}\\
\midrule
\multirow{5}{*}{UniDA}
 & UAN\cite{you2019universal}    & 51.6 & 51.7 & 54.3 & 61.7 & 57.6 & 61.9 & 50.4 & 47.6 & 61.5 & 62.9 & 52.6 & 65.2 & 56.6\\
 & DANCE\cite{saito2020universal}  & 61.0 & 60.4 & 64.9 & 65.7 & 58.8 & 61.8 & 73.1 & 61.2 & 66.6 & 67.7 & 62.4 & 63.7 & 63.9\\
 & DCC\cite{li2021domain}    & 58.0 & 54.1 & 58.0 & 74.6 & 70.6 & 77.5 & 64.3 & \emph{73.6} & 74.9 & 81.0 & \emph{75.1} & 80.4 & 70.2\\
 & OVANet\cite{saito2021ovanet} & 61.7 & 77.3 & 77.9 & 69.7 & 69.2 & 75.7 & 70.6 & 58.2 & 79.2 & 76.8 & 63.0 & 79.0 & 71.6\\
 & CCCOP\cite{zhou2025cross}  & \emph{69.8} & \emph{88.9} & \emph{90.0} & \emph{84.3} & \emph{85.6} & \emph{88.2} & \emph{79.3} & 70.6 & \emph{89.5} & \emph{84.3} & 72.0 & \emph{86.5} & \emph{82.5}\\
\midrule
\multirow{5}{*}{SF-UniDA} 
 & UMAD\cite{liang2021umad}     & 58.7 & 77.1 & 85.4 & 70.0 & 67.6 & 76.3 & 69.0 & 56.8 & 83.0 & 79.5 & 62.0 & 80.8 & 72.2\\
 & LEAD\cite{qu2024lead}     & 62.7 & 78.1 & 86.4 & 70.6 & 76.3 & 83.4 & 75.3 & 60.6 & 86.2 & 75.4 & 60.7 & 83.7 & 75.0\\
 & GLC++\cite{qu2025glc++}    & 69.0 & 78.8 & 90.0 & 70.1 & 81.2 & 86.1 & \underline{78.8} & 57.9 & 89.2 & \underline{81.3} & 54.2 & \underline{87.0} & 77.0\\
 & EGRS\cite{lu2026exploring}     & \underline{74.5} & \underline{80.8} & \underline{90.2} & 71.6 & \underline{83.1} & \underline{90.1} & 77.7 & 54.8 & \underline{89.8} & 81.0 & 63.6 & 86.5 & \underline{78.6}\\
 & ESBN\cite{li2025esbn}     & 64.7 & 78.4 & 89.0 & \underline{71.7} & 74.0 & 80.0 & 76.5 & \underline{62.0} & 87.5 & 76.5 & \underline{65.1} & 82.4 & 75.7\\
\midrule
\multirow{5}{*}{B$^2$-UniDA}
 & SO         & 48.8 & 64.1 & 77.1 & 48.6 & 62.0 & 60.8 & 44.0 & 31.9 & 66.9 & 65.4 & 45.0 & 74.6 & 57.4\\
 & UB2DA\cite{deng2021universal}      & 60.9 & 69.6 & 76.3 & 74.4 & 69.2 & 76.5 & 74.5 & 60.3 & 76.2 & 74.1 & 62.0 & 71.1 & 70.4\\
 & ADU\cite{lai2025adu}        & 61.2 & 72.7 & 77.9 & 70.3 & 72.5 & 77.3 & 75.9 & 62.0 & 84.7 & 73.2 & 64.1 & 74.9 & 72.2\\
 & GSS\cite{he2025gradient}        & \textbf{67.9} & \textbf{83.2} & \textbf{86.4} & 76.4 & 76.6 & \textbf{85.5} & 77.6 & 65.2 & \textbf{87.0} & 78.7 & 67.6 & \textbf{82.1} & \textbf{77.9}\\
 & PReS(ours) & 66.5 & 75.2 & 84.2 & \textbf{82.4} & \textbf{77.9} & 84.0 & \textbf{82.8} & \textbf{67.7} & 84.8 & \textbf{81.5} & \textbf{68.2} & 75.5 & 77.6\\
\bottomrule
\end{tabular}
\end{adjustbox}
\label{officehomeop}
\end{table*}

\begin{table*}[htb]
\centering
\caption{H-score (\%) comparison under OSDA on Office-Home, Office-31, and VisDA. Italic, underlined, and bold denote the best UniDA, SF-UniDA, and B$^2$-UniDA results, respectively.}
\begin{adjustbox}{max width=\textwidth}
\begin{tabular}{c | c | c c c c c c c c c c c c c | c | c }
\toprule
\multirow{2}{*}{\textbf{Type}} & \multirow{2}{*}{\textbf{Method}} & \multicolumn{13}{c|}{\textbf{Office-Home (25/0/40)}} & \textbf{Office-31 (10/0/11)} & \textbf{VisDA (6/0/6)} \\
\cmidrule(lr){3-15}
\cmidrule(lr){16-16}
\cmidrule(lr){17-17}
& & Ar$\rightarrow$Cl & Ar$\rightarrow$Pr & Ar$\rightarrow$Rw & Cl$\rightarrow$Ar & Cl$\rightarrow$Pr & Cl$\rightarrow$Rw & Pr$\rightarrow$Ar & Pr$\rightarrow$Cl &  Pr$\rightarrow$Rw & Rw$\rightarrow$Ar & Rw$\rightarrow$Cl & Rw$\rightarrow$Pr & Avg. & Avg. & Avg. \\
\midrule
\multirow{5}{*}{UniDA}
 & UAN\cite{you2019universal}    & 34.7 & 22.4 & 9.4 & 38.9 & 22.9 & 21.8 & 47.4 & 39.7 & 30.9 & 34.4 & 35.8 & 22.0 & 30.0 & 55.1 & 51.9\\
 & DANCE\cite{saito2020universal}  & 6.5 & 9.0 & 9.9 & 20.4 & 10.4 & 9.2 & 28.4 & 12.8 & 12.6 & 14.2 & 7.9 & 13.2 & 12.9 & 79.8 & 67.5\\
 & DCC\cite{li2021domain}    & 56.1 & 67.5 & 66.7 & 49.6 & 66.5 & 64.0 & 55.8 & 53.0 & 70.5 & 61.6 & 57.2 & 71.9 & 61.7 & 72.7 & 59.6\\
 & OVANet\cite{saito2021ovanet} & 59.3 & 66.9 & 68.1 & 61.0 & 64.3 & 66.4 & 60.5 & 53.9 & 68.0 & 68.2 & 59.6 & 65.4 & 63.4 & 90.0 & 58.4\\
 & CCCOP\cite{zhou2025cross}  & \emph{66.9} & \emph{74.6} & \emph{78.5} & \emph{69.5} & \emph{78.8} & \emph{80.2} & \emph{70.1} & \emph{62.2} & \emph{79.3} & \emph{71.9} & \emph{64.4} & \emph{77.9} & \emph{71.8} & \emph{92.5} & \emph{75.2}\\
\midrule
\multirow{5}{*}{SF-UniDA} 
 & UMAD\cite{liang2021umad}     & 53.6 & 63.4 & 73.1 & 54.3 & 64.7 & 66.7 & 58.2 & 53.5 & 71.0 & 67.3 & 54.2 & 72.5 & 62.7 & \underline{92.2} & 66.8\\
 & LEAD\cite{qu2024lead}     & 60.7 & 70.8 & 76.5 & 61.0 & 68.6 & 70.8 & 65.5 & 59.8 & 74.2 & 64.8 & 57.7 & 75.8 & 67.2 & 90.3 & \underline{74.2}\\
 & GLC++\cite{qu2025glc++}    & 65.2 & 72.9 & \underline{78.6} & \underline{61.7} & 71.9 & 74.2 & 63.2 & \underline{65.3} & 76.5 & \underline{67.4} & \underline{65.9} & 77.7 & \underline{70.1} & 91.4 & 73.1\\
 & EGRS\cite{lu2026exploring}     & \underline{67.2} & 72.0 & \underline{78.6} & 60.3 & 70.3 & \underline{75.4} & 63.7 & 64.0 & 76.9 & 66.1 & 64.6 & \underline{78.3} & 69.8 & 89.3 & 73.3\\
 & ESBN\cite{li2025esbn}     & 59.7 & \underline{79.0} & 72.6 & 57.6 & \underline{75.0} & 69.6 & \underline{65.6} & 50.2 & \underline{77.8} & 62.8 & 57.4 & 77.1 & 67.0 & 91.8 & 73.6\\
\midrule
\multirow{4}{*}{B$^2$-UniDA}
 & SO         & 48.1 & 59.4 & 73.2 & 41.8 & 53.5 & 56.0 & 40.0 & 29.9 & 63.4 & 61.8 & 45.3 & 69.9 & 53.5 & 57.0 & 41.3\\
 & UB2DA\cite{deng2021universal}      & 65.5 & 70.4 & 75.5 & 67.8 & 69.3 & 74.4 & \textbf{71.2} & 56.7 & 75.0 & 70.7 & 63.3 & 69.8 & 69.1 & 87.1 & 48.1\\
 & ADU\cite{lai2025adu}        & \textbf{66.0} & 70.5 & 77.4 & \textbf{72.2} & 70.1 & 75.0 & 69.0 & \textbf{63.6} & 76.1 & \textbf{73.4} & \textbf{64.1} & \textbf{75.2} & 71.1 & 88.1 & 50.8\\
 & PReS(ours) & 63.8 & \textbf{73.1} & \textbf{78.1} & 70.8 & \textbf{72.6} & \textbf{76.9} & \textbf{71.2} & 62.3 & \textbf{79.5} & 72.5 & 62.2 & 73.5 & \textbf{71.4} & \textbf{90.8} & \textbf{69.2}\\
\bottomrule
\end{tabular}
\end{adjustbox}
\label{office31os}
\end{table*}

\begin{table*}[htb]
\centering
\caption{Accuracy (\%) comparison under PDA on Office-Home, Office-31, and VisDA. Italic, underlined, and bold denote the best UniDA, SF-UniDA, and B$^2$-UniDA results, respectively.}
\begin{adjustbox}{max width=\textwidth}
\begin{tabular}{c | c | c c c c c c c c c c c c c | c | c }
\toprule
\multirow{2}{*}{\textbf{Type}} & \multirow{2}{*}{\textbf{Method}} & \multicolumn{13}{c|}{\textbf{Office-Home (25/40/0)}} & \textbf{Office-31 (10/21/0)} & \textbf{VisDA (6/6/0)} \\
\cmidrule(lr){3-15}
\cmidrule(lr){16-16}
\cmidrule(lr){17-17}
& & Ar$\rightarrow$Cl & Ar$\rightarrow$Pr & Ar$\rightarrow$Rw & Cl$\rightarrow$Ar & Cl$\rightarrow$Pr & Cl$\rightarrow$Rw & Pr$\rightarrow$Ar & Pr$\rightarrow$Cl &  Pr$\rightarrow$Rw & Rw$\rightarrow$Ar & Rw$\rightarrow$Cl & Rw$\rightarrow$Pr & Avg. & Avg. & Avg. \\
\midrule
\multirow{4}{*}{UniDA}
 & DANCE\cite{saito2020universal}  & 53.6 & 73.2 & 84.9 & 70.8 & 67.3 & 82.6 & 70.0 & 50.9 & 84.8 & 77.0 & 55.9 & 81.8 & 71.1 & 86.0 & 73.7\\
 & DCC\cite{li2021domain}    & 54.2 & 47.5 & 57.5 & 83.8 & 71.6 & 86.2 & 63.7 & 65.0 & 75.2 & \emph{85.5} & \emph{78.2} & 82.6 & 70.9 & 93.3 & 72.4\\
 & OVANet\cite{saito2021ovanet} & 50.9 & 71.2 & 80.2 & 60.9 & 61.1 & 69.3 & 61.1 & 45.4 & 77.6 & 72.2 & 52.7 & 77.6 & 65.0 & 89.4 & 58.5\\
 & CCCOP\cite{zhou2025cross}  & \emph{66.7} & \emph{84.5} & \emph{88.4} & \emph{84.9} & \emph{82.0} & \emph{87.8} & \emph{73.9} & \emph{66.2} & \emph{87.9} & 79.3 & 69.8 & \emph{85.6} & \emph{79.8} & \emph{96.8} & \emph{75.1}\\
\midrule
\multirow{5}{*}{SF-UniDA} 
& UMAD\cite{liang2021umad}     & 51.2 & 66.5 & 79.2 & 63.1 & 62.9 & 68.2 & 63.3 & 56.4 & 75.9 & 74.5 & 55.9 & 78.3 & 66.3 & 89.5 & 68.5\\
& LEAD\cite{qu2024lead}     & 58.2 & \underline{83.1} & 87.0 & 70.5 & \underline{75.4} & 83.3 & 73.7 & 50.4 & \underline{83.7} & 78.3 & 58.7 & 83.2 & 73.8 & \underline{95.5} & 75.3\\
& GLC++\cite{qu2025glc++}    & 58.1 & 79.6 & 89.3 & 74.8 & 72.2 & 80.6 & \underline{76.6} & 42.0 & 83.2 & 78.5 & 60.8 & 85.2 & 73.4 & 94.1 & 75.8\\
& EGRS\cite{lu2026exploring}     & \underline{58.7} & 78.2 & \underline{89.5} & \underline{75.4} & 73.1 & 81.8 & 76.4 & \underline{61.7} & 83.4 & \underline{79.2} & \underline{62.9} & \underline{85.6} & \underline{75.5} & 94.0 & 75.7\\
& ESBN\cite{li2025esbn}     & 56.0 & 76.6 & 86.5 & 71.8 & 71.4 & \underline{84.3} & 73.2 & 45.6 & 83.6 & \underline{80.6} & 52.0 & 84.6 & 72.2 & \underline{95.5} & \underline{76.2}\\
\midrule
\multirow{2}{*}{B$^2$-UniDA}
 & SO         & 45.5 & 69.0 & 80.8 & 56.5 & 58.8 & 66.8 & 61.6 & 37.6 & 76.9 & 72.8 & 46.4 & 79.5 & 62.7 & 87.2 & 58.6\\
 & PReS(ours) & \textbf{63.2} & \textbf{88.0} & \textbf{88.4} & \textbf{78.3} & \textbf{84.5} & \textbf{85.9} & \textbf{80.6} & \textbf{61.7} & \textbf{87.9} & \textbf{79.8} & \textbf{62.8} & \textbf{88.8} & \textbf{79.2} & \textbf{96.4} & \textbf{80.3}\\
\bottomrule
\end{tabular}
\end{adjustbox}
\label{office31pda}
\end{table*}

\begin{table*}[t]
\centering
\caption{Accuracy (\%) comparison under CDA on Office-Home and Office-31. Italic, underlined, and bold denote the best UniDA, SF-UniDA, and B$^2$-UniDA results, respectively.}
\begin{adjustbox}{max width=\textwidth}
\begin{tabular}{c | c | c c c c c c c c c c c c c | c }
\toprule
\multirow{2}{*}{\textbf{Type}} & \multirow{2}{*}{\textbf{Method}} & \multicolumn{13}{c|}{\textbf{Office-Home (65/0/0)}} & \textbf{Office-31 (31/0/0)} \\
\cmidrule(lr){3-15}
\cmidrule(lr){16-16}
& & Ar$\rightarrow$Cl & Ar$\rightarrow$Pr & Ar$\rightarrow$Rw & Cl$\rightarrow$Ar & Cl$\rightarrow$Pr & Cl$\rightarrow$Rw & Pr$\rightarrow$Ar & Pr$\rightarrow$Cl &  Pr$\rightarrow$Rw & Rw$\rightarrow$Ar & Rw$\rightarrow$Cl & Rw$\rightarrow$Pr & Avg. & Avg. \\
\midrule
\multirow{5}{*}{UniDA}
 & UAN\cite{you2019universal}    & 45.0 & 63.6 & 71.2 & 51.4 & 58.2 & 63.2 & 52.6 & 40.9 & 71.0 & 63.3 & 48.2 & 75.4 & 58.7 & 84.4\\
 & DANCE\cite{saito2020universal}  & 54.3 & 75.9 & 78.4 & 64.8 & 72.1 & 73.4 & 63.2 & 53.0 & 79.4 & 73.0 & 58.2 & 82.9 & 69.1 & 85.5\\
 & DCC\cite{li2021domain}    & 35.4 & 61.4 & 75.2 & 45.7 & 59.1 & 62.7 & 43.9 & 30.9 & 70.2 & 57.8 & 41.0 & 77.9 & 55.1 & 87.4\\
 & OVANet\cite{saito2021ovanet} & 48.6 & 68.6 & 76.1 & 58.2 & 66.0 & 68.9 & 55.9 & 44.7 & 75.9 & 68.9 & 52.0 & 79.8 & 63.6 & 82.6\\
 & CCCOP\cite{zhou2025cross}  & \textit{59.8} & \textit{83.1} & \textit{84.8} & \textit{68.1} & \textit{81.8} & \textit{80.7} & \textit{69.7} & \textit{59.6} & \textit{83.4} & \textit{73.7} & \textit{59.9} & \textit{86.6} & \textit{73.8} & \textit{93.2}\\
\midrule
\multirow{3}{*}{SF-UniDA} 
 & UMAD\cite{liang2021umad}     & 47.7 & 64.6 & 73.0 & 58.8 & 64.7 & 68.1 & 58.2 & 47.4 & 74.2 & 69.2 & 52.7 & 77.9 & 63.1 & 81.4\\
& GLC++\cite{qu2025glc++}    & 53.5 & 77.0 & \underline{80.3} & 65.7 & \underline{79.6} & 78.7 & \underline{65.6} & \underline{56.1} & \underline{81.6} & \underline{71.5} & \underline{58.5} & \underline{84.0} & \underline{71.0} & 88.4\\
& EGRS\cite{lu2026exploring}     & \underline{55.3} & \underline{77.5} & 80.1 & \underline{66.4} & 78.3 & \underline{79.2} & 65.3 & 55.8 & 81.4 & 71.2 & 57.7 & 83.5 & \underline{71.0} & \underline{89.0}\\
\midrule
\multirow{2}{*}{B$^2$-UniDA}
 & SO         & 46.0 & 70.5 & 77.4 & 62.2 & 70.1 & 75.0 & 59.0 & 43.6 & 76.1 & 68.4 & 50.1 & 70.2 & 64.1 & 78.1\\
 & PReS(ours) & \textbf{55.5} & \textbf{82.5} & \textbf{83.2} & \textbf{69.1} & \textbf{81.1} & \textbf{81.7} & \textbf{68.9} & \textbf{53.2} & \textbf{82.8} & \textbf{71.0} & \textbf{54.6} & \textbf{84.5} & \textbf{72.3} & \textbf{86.2}\\
\bottomrule
\end{tabular}
\end{adjustbox}
\label{office31cda}
\end{table*}

\subsection{Setup}
\emph{\textbf{Datasets and Adaptation Scenarios}}:
We evaluate PReS on four widely used domain adaptation benchmarks: Office-31~\cite{saenko2010adapting}, Office-Home~\cite{venkateswara2017deep}, VisDA~\cite{peng2017visda}, and DomainNet~\cite{peng2019moment}. \textbf{Office-31} contains 4,652 images from 31 categories across Amazon (A), Webcam (W), and DSLR (D), forming six bidirectional adaptation tasks. \textbf{Office-Home} contains approximately 15,500 images from 65 categories across Art (Ar), Clipart (Cl), Product (Pr), and Real-World (Rw), forming 12 adaptation tasks. \textbf{VisDA} contains over 152,000 synthetic and 55,000 real images from 12 categories, with Synthetic (S) $\rightarrow$ Real (R) as the adaptation task. \textbf{DomainNet} contains approximately 600,000 images from 345 categories across six domains; following previous studies, we use Painting (P), Real (R), and Sketch (S) for evaluation.

We consider four adaptation scenarios with different source-target label-space relationships: open-partial domain adaptation (OPDA), open-set domain adaptation (OSDA), partial domain adaptation (PDA), and closed-set domain adaptation (CDA). Table~\ref{classnum} summarizes the corresponding class splits.

\emph{\textbf{Evaluation Metrics}}:
Following previous UniDA studies~\cite{fu2020learning,he2025gradient}, we report classification accuracy for CDA and PDA, where no target-private classes exist. For OSDA and OPDA, we adopt the H-score to jointly evaluate known-class classification and unknown-class recognition:
\begin{equation}
H = 2\frac{Acc_{kno}\cdot Acc_{unk}}{Acc_{kno}+Acc_{unk}},
\label{eq:hscore}
\end{equation}
where $Acc_{kno}$ denotes the mean per-class accuracy over the shared classes, and $Acc_{unk}$ denotes the accuracy of recognizing target-private samples as unknown.

\emph{\textbf{Comparison Methods}}:
We compare PReS with representative UniDA, SF-UniDA, and B$^{2}$-UniDA methods. The UniDA methods include UAN~\cite{you2019universal}, DANCE~\cite{saito2020universal}, DCC~\cite{li2021domain}, OVANet~\cite{saito2021ovanet}, and CCCOP~\cite{zhou2025cross}; the SF-UniDA methods include UMAD~\cite{liang2021umad}, LEAD~\cite{qu2024lead}, GLC++~\cite{qu2025glc++}, EGRS~\cite{lu2026exploring}, and ESBN~\cite{li2025esbn}; and the B$^{2}$-UniDA methods include UB2DA~\cite{deng2021universal}, ADU~\cite{lai2025adu}, and GSS~\cite{he2025gradient}. The reported results are collected from the original publications or, when applicable, from LEAD~\cite{qu2024lead}, GLC++~\cite{qu2025glc++}, and ADU~\cite{lai2025adu}. Note that ADU considers only OSDA and OPDA rather than the general B$^{2}$-UniDA setting. Since existing B$^{2}$-UniDA studies generally do not report reproducible PDA and CDA results, we additionally use Source-Only (SO), which directly applies the pretrained source model to target samples without adaptation, as the B$^{2}$-UniDA baseline for these settings.

\emph{\textbf{Implementation Details}} \label{detail}:
All experiments are implemented in PyTorch~\cite{paszke2017automatic}. Following previous B$^{2}$-UniDA studies~\cite{he2025gradient}, we use an ImageNet-pretrained ResNet-50~\cite{he2016deep,deng2009imagenet} as the backbone. The source model is trained with cross-entropy loss and subsequently accessed only through black-box queries during adaptation. Both source-model training and target-model adaptation use SGD with a learning rate of $10^{-2}$, momentum of $0.9$, weight decay of $5\times10^{-4}$, and batch size of $64$. We adopt CLIP ViT-B/32 and Stable Diffusion v1.5~\cite{rombach2022high} as the foundation models. The refinement starting epoch $t$ is set to $5$. The hyperparameters $\beta$ and $\gamma$ in Eq.~(\ref{lossunk}) are set to $0.3$ and $0.1$, respectively, except for DomainNet, where they are set to $0.025$ and $0.01$. 
We set $\epsilon=10^{-6}$ in Eq.~(\ref{eq:fusion_weight}) and the uncertainty threshold $\omega=0.55$ in Eq.~(\ref{eq:final_prediction}), following GLC~\cite{qu2023upcycling}.

\subsection{Experimental Results}
\emph{\textbf{Results for OPDA}}:
Tables~\ref{office31op} and~\ref{officehomeop} report the OPDA results on the four benchmarks. PReS achieves the best average H-score among B$^{2}$-UniDA methods on Office-31, VisDA, and DomainNet, with 91.2\%, 66.5\%, and 58.4\%, respectively. In particular, on VisDA and DomainNet, PReS outperforms GSS by 23.0 and 8.1 percentage points and ADU by 17.8 and 7.4 points, respectively. On Office-Home, PReS achieves 77.6\%, only 0.3 points below GSS, while obtaining the best B$^{2}$-UniDA results on 6 of the 12 transfer tasks. Despite the more restrictive source-side access, PReS also remains competitive with UniDA and SF-UniDA methods, e.g., achieving 91.2\% versus the best SF-UniDA result of 91.7\% on Office-31 and 58.4\% versus the best UniDA result of 60.0\% on DomainNet.

\emph{\textbf{Results for OSDA}}:
Table~\ref{office31os} reports the OSDA results. PReS achieves the best performance among B$^{2}$-UniDA methods on all three benchmarks, with H-scores of 71.4\%, 90.8\%, and 69.2\% on Office-Home, Office-31, and VisDA, respectively. Compared with ADU, these results correspond to improvements of 0.3, 2.7, and 18.4 percentage points. PReS is also competitive with methods having stronger source-side access, outperforming all compared SF-UniDA methods on Office-Home and achieving results close to the best UniDA/SF-UniDA methods on Office-31.

\emph{\textbf{Results for PDA}}:
Table~\ref{office31pda} reports the PDA results. PReS achieves accuracies of 79.2\%, 96.4\%, and 80.3\% on Office-Home, Office-31, and VisDA, respectively, improving over SO by 16.5, 9.2, and 21.7 percentage points. It outperforms all compared SF-UniDA methods on Office-Home and Office-31 and is within 0.6 and 0.4 points of the best UniDA results, respectively. On VisDA, PReS surpasses all compared UniDA and SF-UniDA methods, exceeding their best results by 5.2 and 4.1 points, respectively.

\emph{\textbf{Results for CDA}}:
As shown in Table~\ref{office31cda}, PReS remains effective when the source and target label spaces are identical. It achieves accuracies of 72.3\% and 86.2\% on Office-Home and Office-31, improving over SO by 8.2 and 8.1 percentage points, respectively. On Office-Home, PReS also outperforms all compared SF-UniDA methods and approaches the best UniDA result (72.3\% vs. 73.8\%).

Overall, PReS performs consistently across diverse label shifts under B$^2$-UniDA and remains competitive with methods accessing source data or source-model parameters.

\subsection{Ablation Study}

\begin{table}[!t]
\centering
\caption{Ablation results of Reliable Known Sample Discovery under OPDA and OSDA. ``Conv. Aug.'' replaces diffusion-based semantic-preserving perturbations with conventional augmentations, while ``w/o CCV'' removes Cross-model Consensus Verification.}
\begin{adjustbox}{max width=0.48\textwidth}
\begin{tabular}{c | c | c c c c}
\toprule
Scenario & Method & Office-31 & Office-Home & VisDA & DomainNet \\
\midrule
\multirow{3}{*}{OSDA}
 & PReS                 & 90.8 & 71.4 & 69.2 & - \\
 \cmidrule(lr){2-2}
 & PReS w/ Conv. Aug.   & 90.3 & 71.0 & 68.3 & - \\
 \cmidrule(lr){2-2}
 & PReS w/o CCV         & 88.6 & 68.2 & 68.6 & - \\
\midrule
\multirow{3}{*}{OPDA}
 & PReS                 & 91.2 & 77.6 & 66.5 & 58.4 \\
 \cmidrule(lr){2-2}
 & PReS w/ Conv. Aug.   & 90.4 & 77.4 & 64.9 & 57.4 \\
 \cmidrule(lr){2-2}
 & PReS w/o CCV         & 90.1 & 75.5 & 65.3 & 54.0 \\
\bottomrule
\end{tabular}
\end{adjustbox}
\label{dif}
\end{table}

\begin{figure}[!t]
\centering
\includegraphics[width=0.46\textwidth]{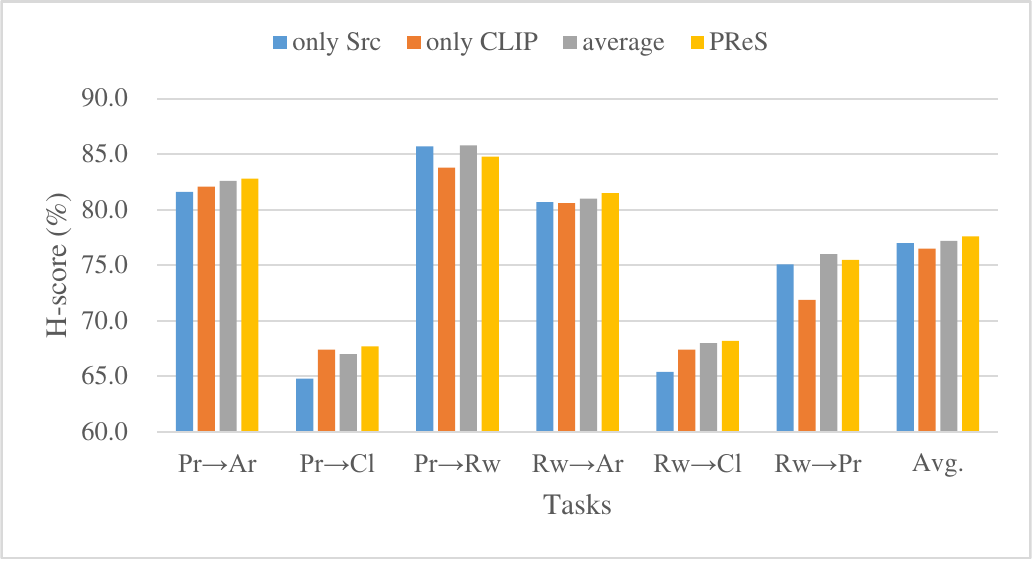}
\caption{Ablation results of Reliability-aware Pseudo-label Generation on Office-Home under OPDA. ``only Src'' and ``only CLIP'' use predictions from the black-box source model and CLIP alone, respectively, while ``average'' equally averages their predictions.}
\label{apfmodule}
\end{figure}

\begin{table}[!h]
\centering
\caption{Ablation results of Asymmetric Target Learning under OPDA. Values denote the average H-score (\%) on each dataset. ‘\(\checkmark\)’ and ‘–’ indicate that the corresponding loss is included and excluded, respectively.}
\begin{adjustbox}{max width=\textwidth}
\begin{tabular}{c c c c | c c c}
\toprule
$\mathcal{L}_{kd}$ & $\mathcal{L}_{im}$ & $\mathcal{L}_{em}$ & $\mathcal{L}_{mix}$ & Office-31 & Office-Home & VisDA \\
\midrule
- & \checkmark & \checkmark & \checkmark & 5.3 & 13.3 & 7.0 \\
\checkmark & - & \checkmark & \checkmark & 77.4 & 67.3 & 60.5 \\
\checkmark & \checkmark & - & \checkmark & 90.9 & 72.7 & 64.1 \\
\checkmark & \checkmark & \checkmark & - & 90.3 & 76.4 & 54.0 \\
\checkmark & \checkmark & \checkmark & \checkmark & 91.2 & 77.6 & 66.5 \\
\bottomrule
\end{tabular}
\end{adjustbox}
\label{ablation}
\end{table}

\begin{figure}[!t]
\centering
\includegraphics[width=0.46\textwidth]{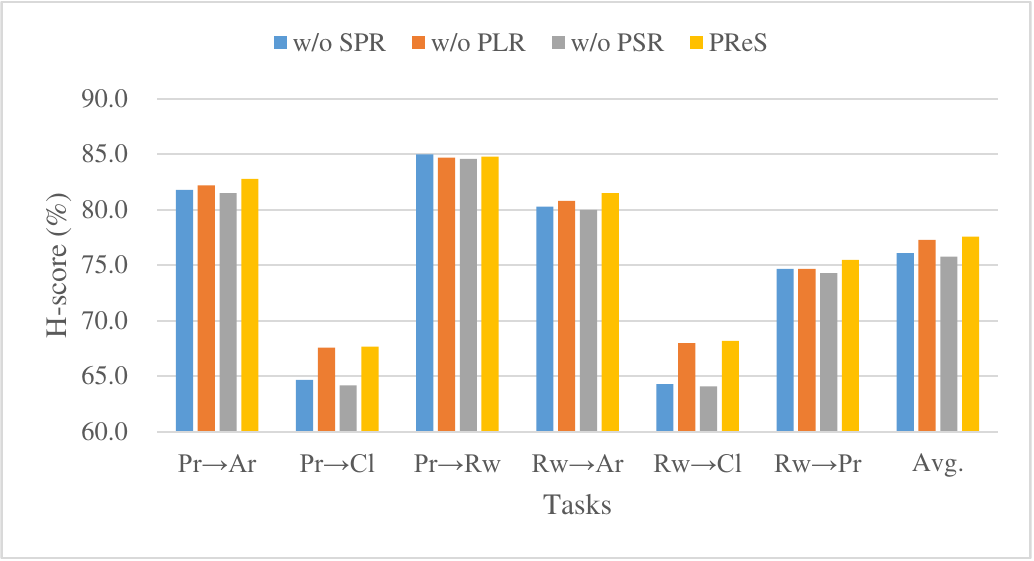}
\caption{Ablation results of Progressive Supervision Refinement (PSR) on Office-Home under OPDA. ``w/o SPR'' and ``w/o PLR'' disable SPR and PLR, respectively, while ``w/o PSR'' disables both.}
\label{mem_model}
\end{figure}

\subsubsection{Ablation on Reliable Known Sample Discovery}
We first evaluate the two successive verification mechanisms in Reliable Known Sample Discovery. For Diffusion-based Semantic Consistency Verification, we replace the diffusion-generated semantic-preserving perturbations with conventional augmentations (Conv. Aug.), including random scaling, horizontal flipping, and random cropping. As shown in Table~\ref{dif}, PReS consistently outperforms this variant under both OPDA and OSDA. Under OPDA, the improvements are 0.8, 0.2, 1.6, and 1.0 percentage points on Office-31, Office-Home, VisDA, and DomainNet, respectively. This indicates that diffusion-based perturbations provide more informative semantic consistency evidence than conventional image transformations, benefiting from their ability to introduce diverse appearance variations while preserving semantic content.

We further remove Cross-model Consensus Verification (CCV) while keeping the remaining components unchanged. This consistently degrades performance, with reductions of 1.1--4.4 points under OPDA and 0.6--3.2 points under OSDA. These results demonstrate that the generalizable semantic knowledge of CLIP provides complementary reliability evidence to the task-specific but source-biased black-box predictions, enabling more reliable known-sample discovery.

\subsubsection{Ablation on Reliability-aware Pseudo-label Generation}
We compare PReS with three alternative pseudo-label generation strategies: using only the black-box source predictions (\emph{only Src}), using only CLIP predictions (\emph{only CLIP}), and equally averaging the two predictions (\emph{average}). As shown in Fig.~\ref{apfmodule}, the two predictors exhibit complementary strengths across transfer tasks, while simple averaging improves the average H-score to 77.2\%, compared with 77.0\% and 76.5\% for \emph{only Src} and \emph{only CLIP}, respectively. By further accounting for their sample-wise reliability, PReS achieves the highest average H-score of 77.6\%. These results validate the complementarity of the two knowledge sources and the effectiveness of reliability-aware fusion.

\subsubsection{Ablation on Asymmetric Target Learning}
To evaluate the learning objectives in Asymmetric Target Learning, we individually remove each loss term under OPDA. As shown in Table~\ref{ablation}, removing $\mathcal{L}_{kd}$ causes the largest performance degradation, confirming the importance of transferring the constructed label-level supervision to the target model. Removing $\mathcal{L}_{im}$ reduces the H-score by 13.8, 10.3, and 6.0 points on Office-31, Office-Home, and VisDA, respectively, demonstrating the benefit of jointly promoting prediction confidence and class diversity. Both $\mathcal{L}_{em}$ and $\mathcal{L}_{mix}$ also contribute to unknown-sample learning, with the latter yielding a particularly large gain of 12.5 points on VisDA. Overall, these results support the asymmetric learning principle of encouraging confident and discriminative predictions for reliable known samples while maintaining uncertain yet locally consistent predictions for potential unknown samples.

\subsubsection{Ablation on Progressive Supervision Refinement}
We evaluate Pseudo-label Refinement (PLR) and Sample Partition Refinement (SPR) using three variants: \emph{w/o SPR}, \emph{w/o PLR}, and \emph{w/o PSR}, where the last disables both refinement mechanisms. As shown in Fig.~\ref{mem_model}, enabling PLR alone improves the average H-score from 75.8\% to 76.1\%, while enabling SPR alone improves it to 77.3\%. Jointly employing PLR and SPR further increases the H-score to 77.6\%. These results demonstrate that both mechanisms contribute to progressive supervision refinement and are complementary when used jointly.

\subsection{Visualization}

\subsubsection{Feature Visualization}

\begin{figure*}[t!]
    \centering 
    \subfigure[SO (D$\rightarrow$A)]{
        \includegraphics[width=0.22\linewidth]{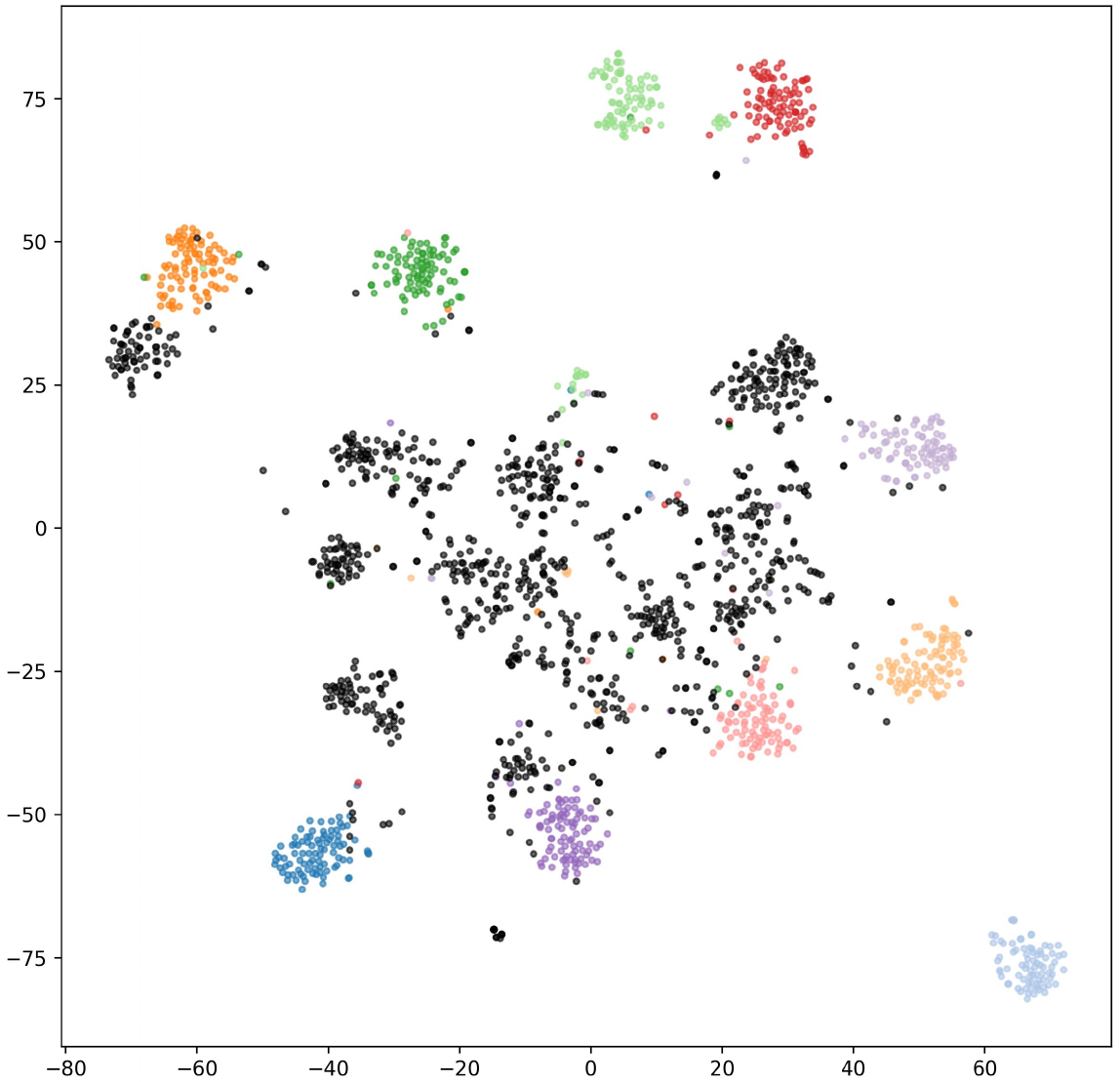} 
    }%
    \subfigure[PReS (D$\rightarrow$A)]{
        \includegraphics[width=0.22\linewidth]{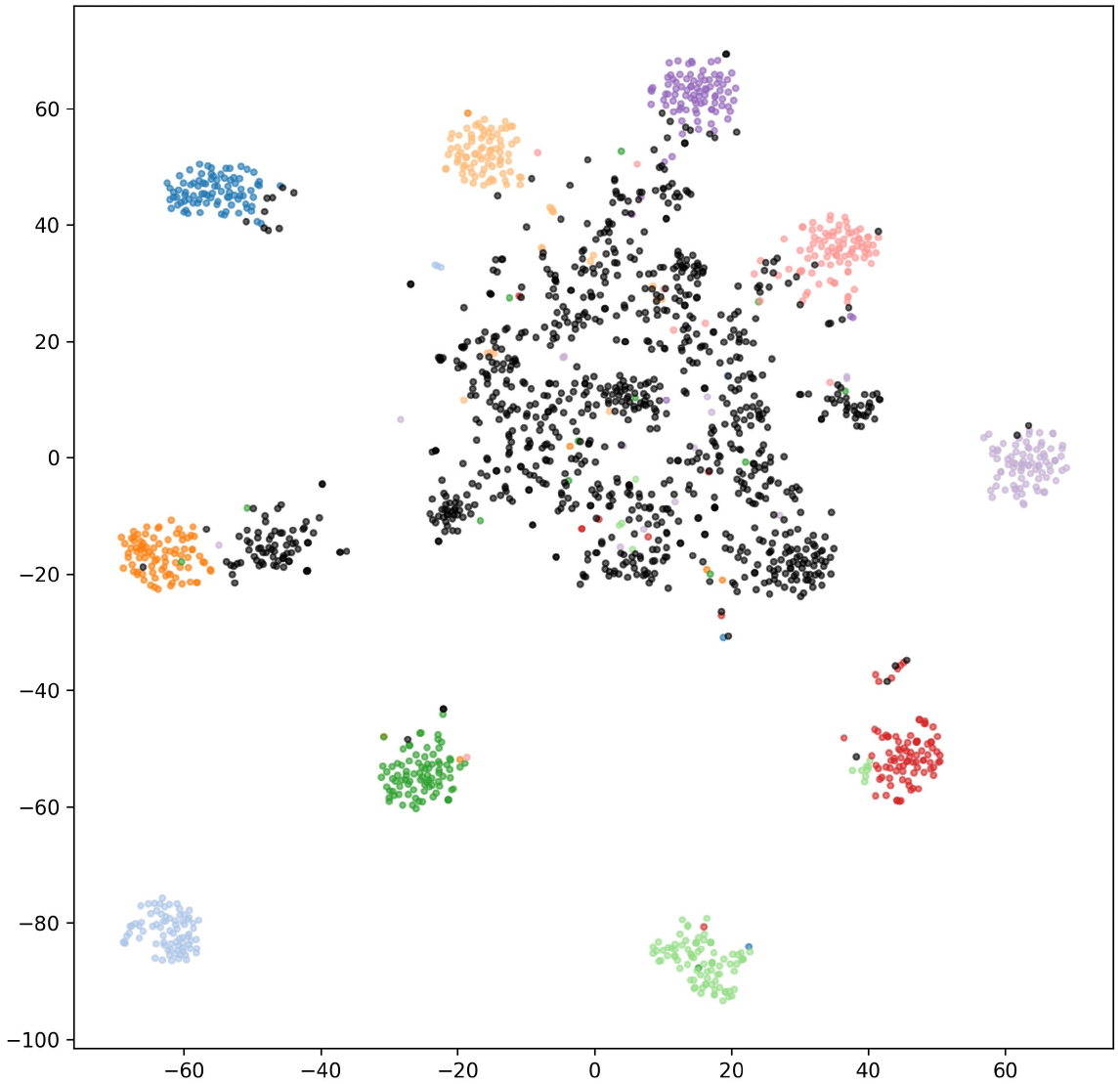}
    }%
    \subfigure[SO (Pr$\rightarrow$Rw)]{
        \includegraphics[width=0.22\linewidth]{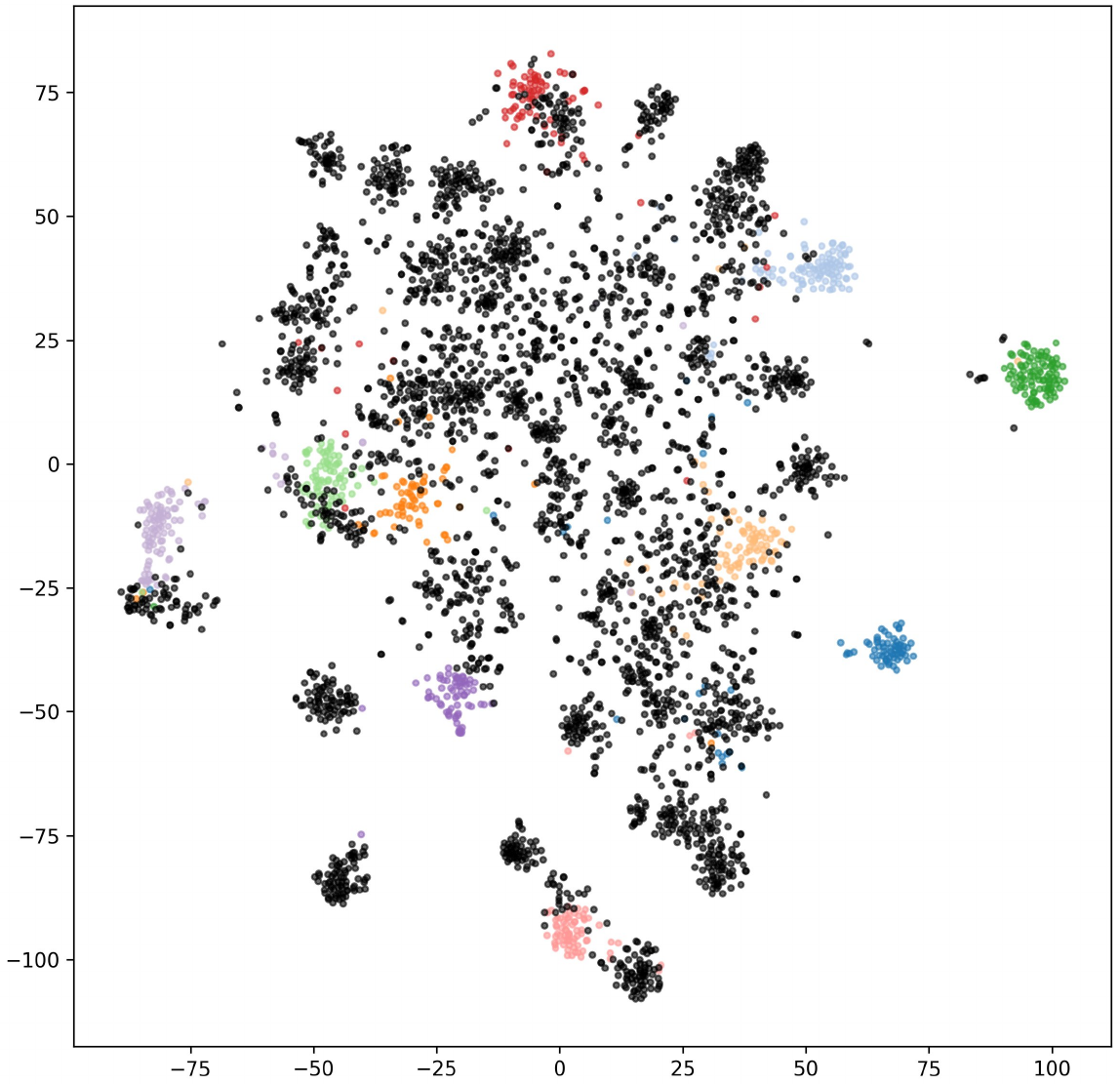}
    }%
    \subfigure[PReS (Pr$\rightarrow$Rw)]{
        \includegraphics[width=0.22\linewidth]{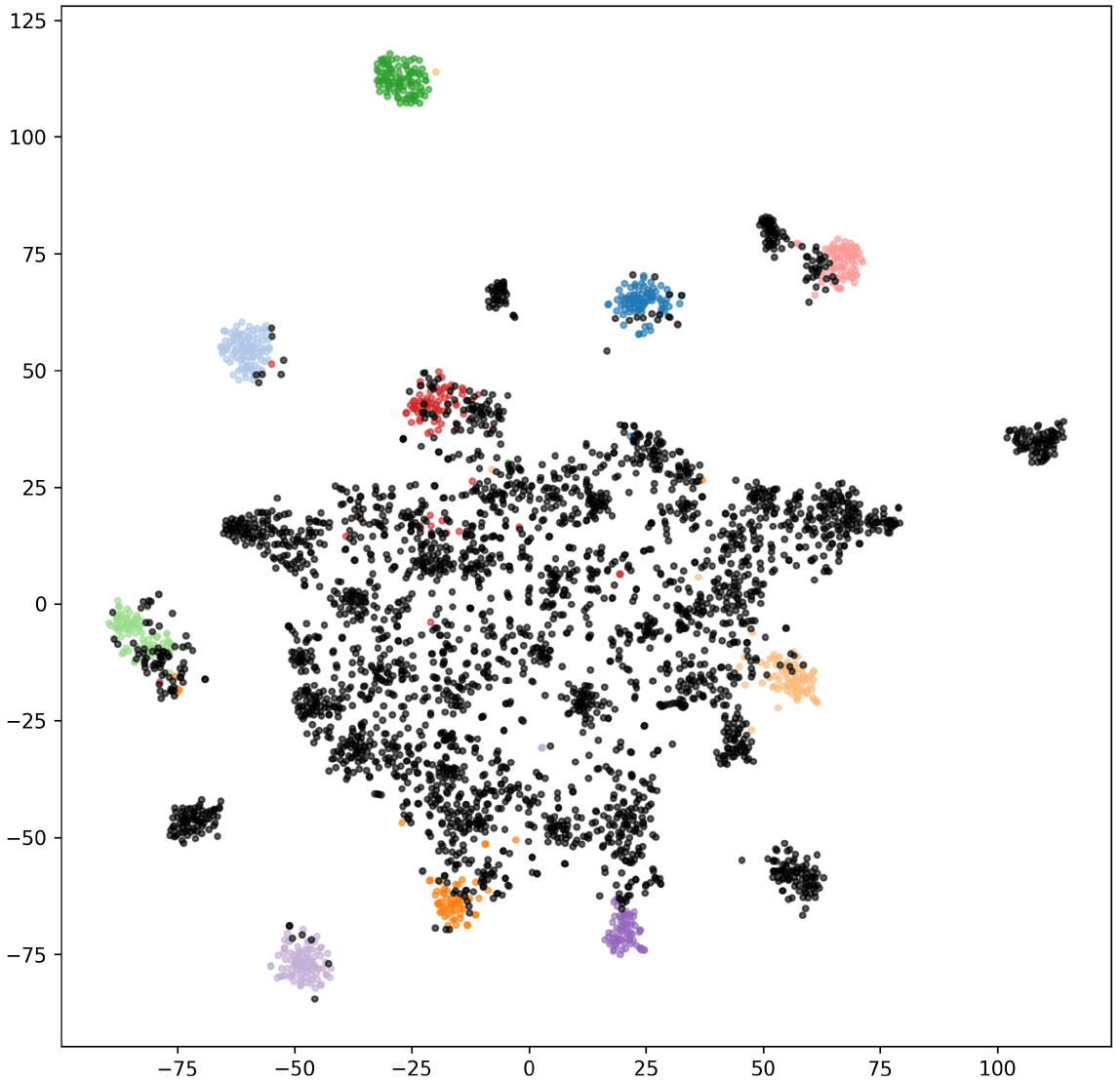}
    }%
    \caption{t-SNE visualization of target features learned by SO and PReS for D$\rightarrow$A on Office-31 and Pr$\rightarrow$Rw on Office-Home under OPDA. Colored and black points represent samples from the shared and target-private classes, respectively. (Best viewed in color and zoomed in.)}
    \label{sne} 
\end{figure*}

Fig.~\ref{sne} presents t-SNE~\cite{van2008visualizing} visualizations of target features learned by the source-only (SO) model and PReS under OPDA. Compared with SO, PReS produces more structured feature distributions, with shared-class samples forming more compact clusters and being more clearly separated from target-private samples. This consistent pattern across both transfer tasks indicates that PReS learns more discriminative target representations under domain and category shifts.

\subsubsection{Discriminative Region Visualization}

\begin{figure}[t!]
    \centering
    \includegraphics[width=0.46\textwidth]{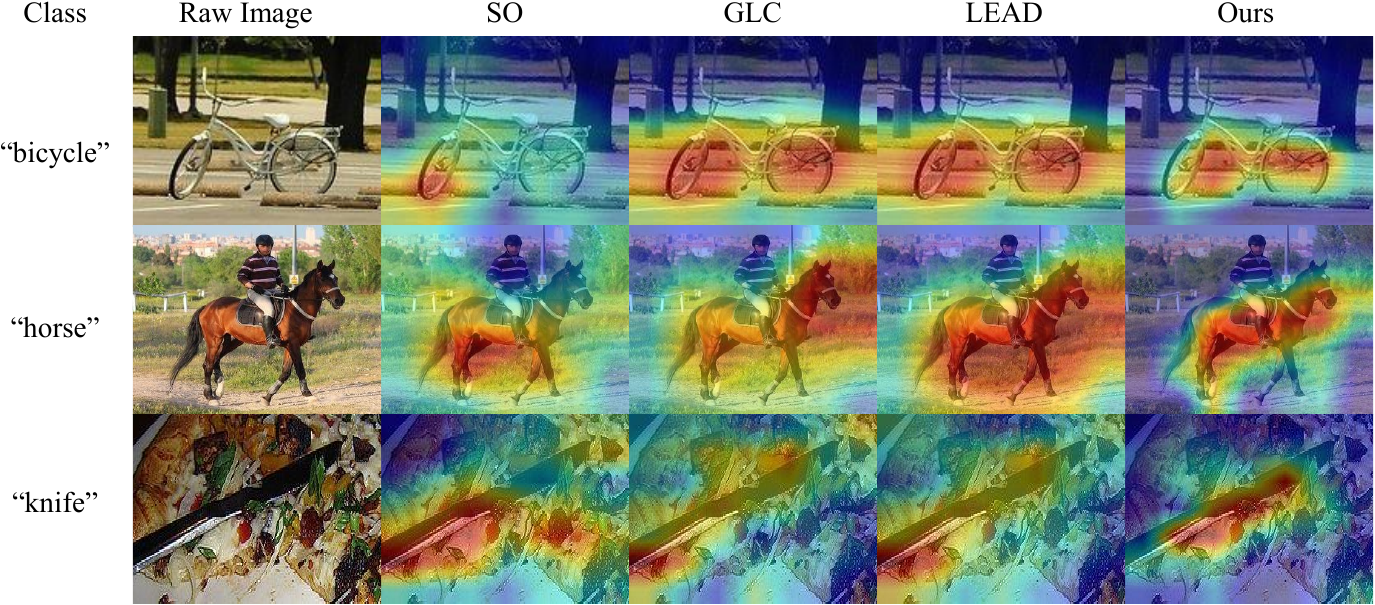}
    \caption{Grad-CAM visualization of discriminative regions on VisDA target samples for the source-only (SO) model, GLC, LEAD, and PReS. Higher activation indicates greater contribution to model predictions.}
    \label{cam}
\end{figure}

Fig.~\ref{cam} presents Grad-CAM~\cite{selvaraju2017grad} visualizations of PReS and three comparison methods on VisDA target samples. Compared with the other methods, PReS generally exhibits more concentrated activation on object-relevant regions and less attention to irrelevant backgrounds, indicating that the adapted target model relies more strongly on semantically relevant visual cues for classification.

\subsection{Parameter Sensitivity}

\begin{figure}[!t]
    \centering
    \begin{minipage}{0.48\linewidth}
        \centering
        \subfigure[Task-wise sensitivity to $\beta$]{
            \includegraphics[width=0.9\linewidth]{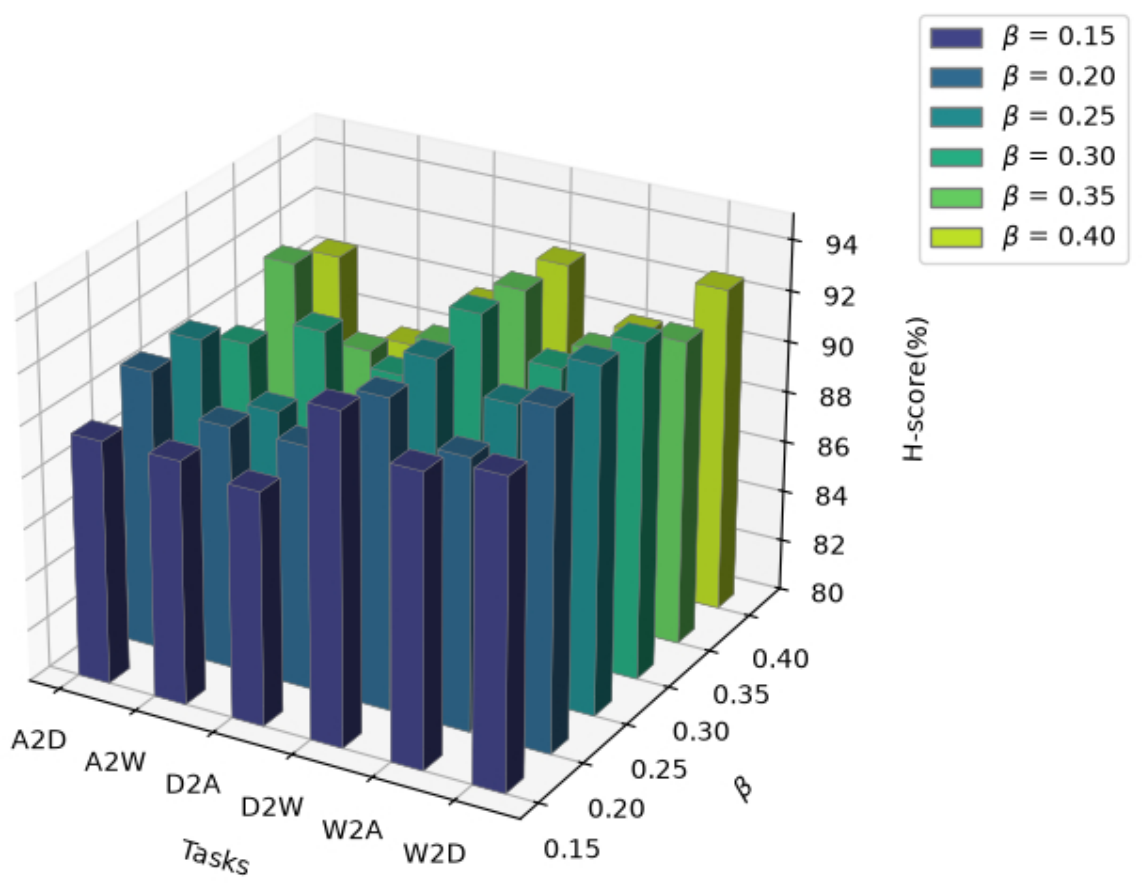}
            \label{con1}
        }
        \hfill
        \subfigure[Average sensitivity to $\beta$]{
            \includegraphics[width=0.9\linewidth]{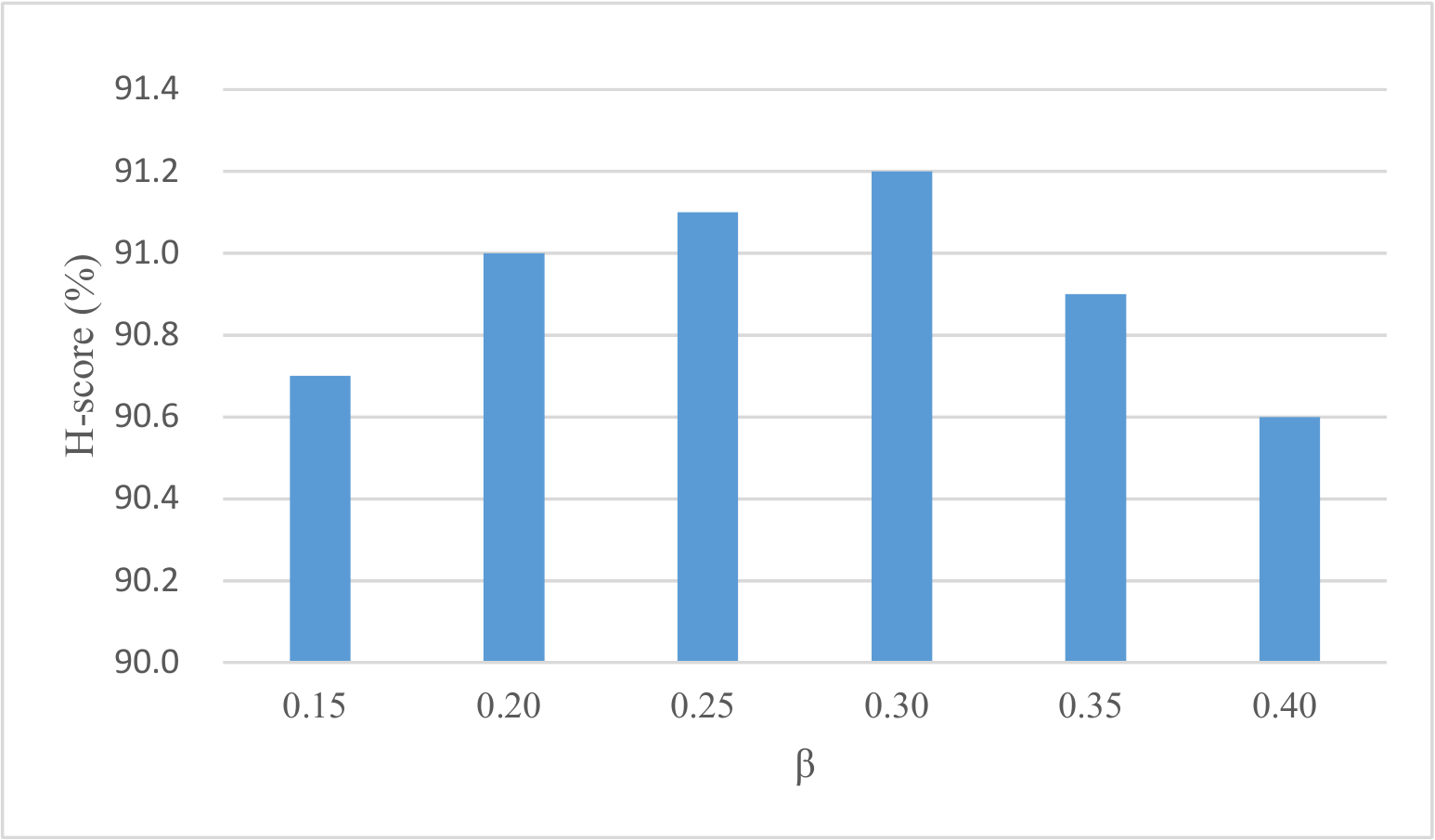}
            \label{con2}
        }
    \end{minipage}
    \hfill
    \begin{minipage}{0.48\linewidth}
        \centering
        \subfigure[Task-wise sensitivity to $\gamma$]{
            \includegraphics[width=0.9\linewidth]{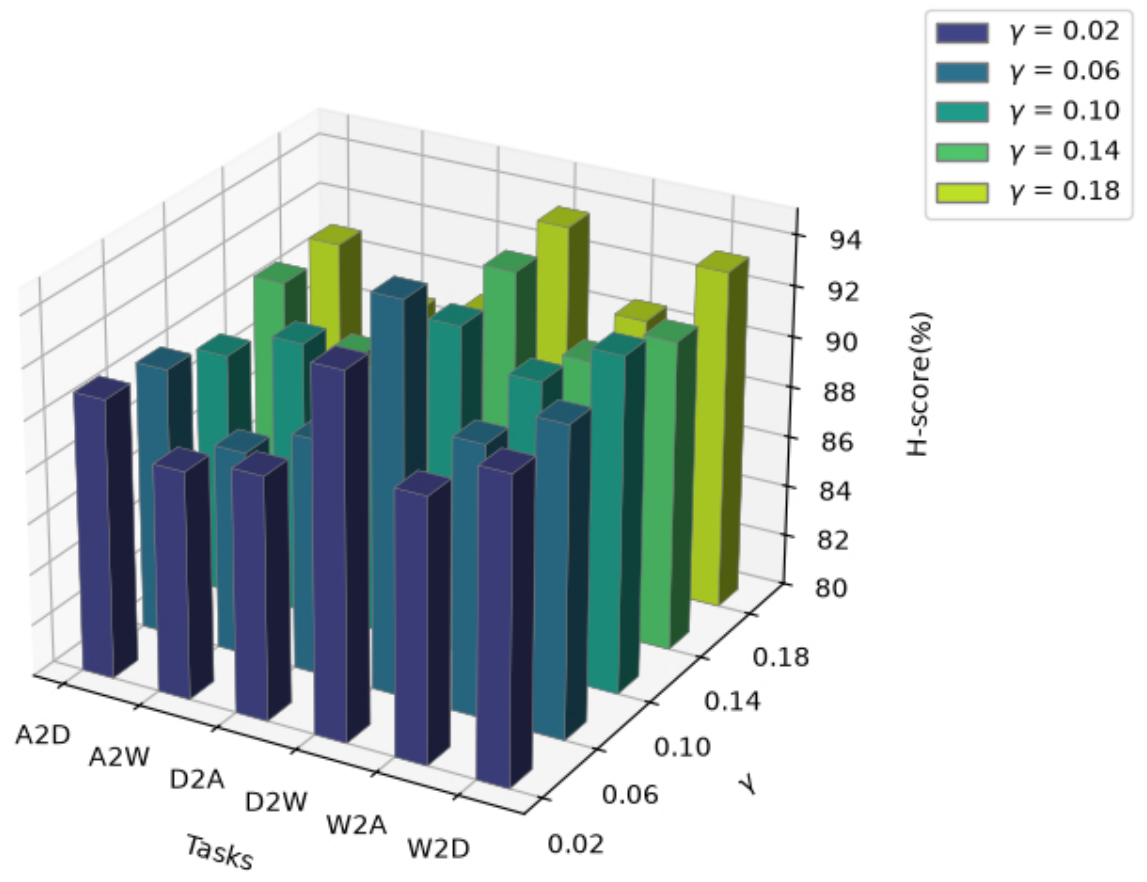}
            \label{con3}
        }
        \hfill
        \subfigure[Average sensitivity to $\gamma$]{
            \includegraphics[width=0.9\linewidth]{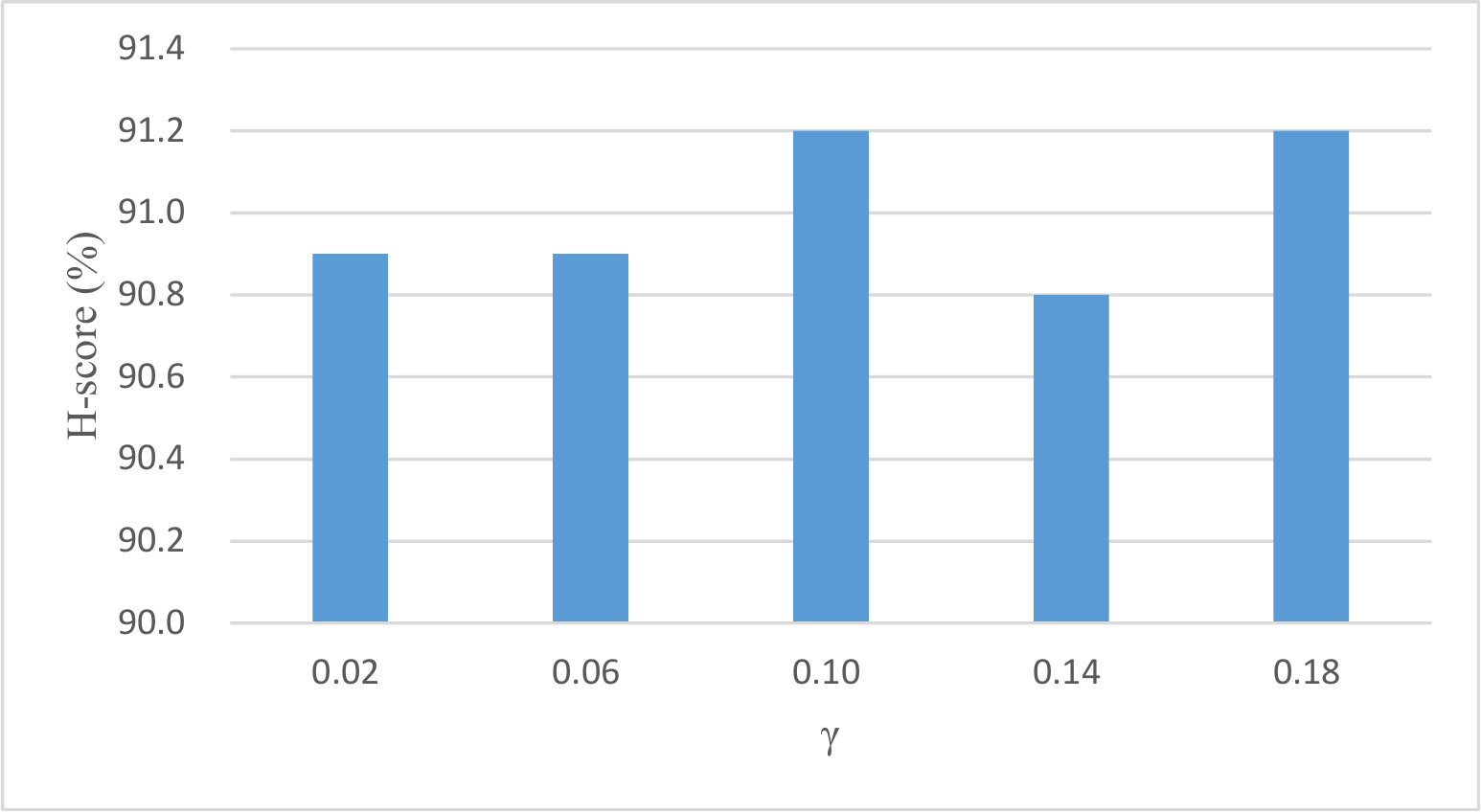}
            \label{con4}
        }
    \end{minipage}
    \caption{Sensitivity of PReS to $\beta$ and $\gamma$ on Office-31 under OPDA. Subfigures (a) and (c) show task-wise H-scores, while (b) and (d) show the corresponding averages.}
    \label{parameter}
\end{figure}

PReS introduces $\beta$ and $\gamma$ in Eq.~(\ref{lossunk}) to balance $\mathcal{L}_{em}$ and $\mathcal{L}_{mix}$, respectively. We vary $\beta$ from $0.15$ to $0.40$ and $\gamma$ from $0.02$ to $0.18$ on Office-31 under OPDA, while fixing the other hyperparameter to its default value. 
As shown in Fig.~6, despite task-wise variations, the average H-score remains stable, ranging from 90.6\% to 91.2\% for $\beta$ and from 90.8\% to 91.2\% for $\gamma$. We therefore set $\beta=0.3$ and $\gamma=0.1$ by default.

\begin{figure}[t!]
\centering
\includegraphics[width=0.46\textwidth]{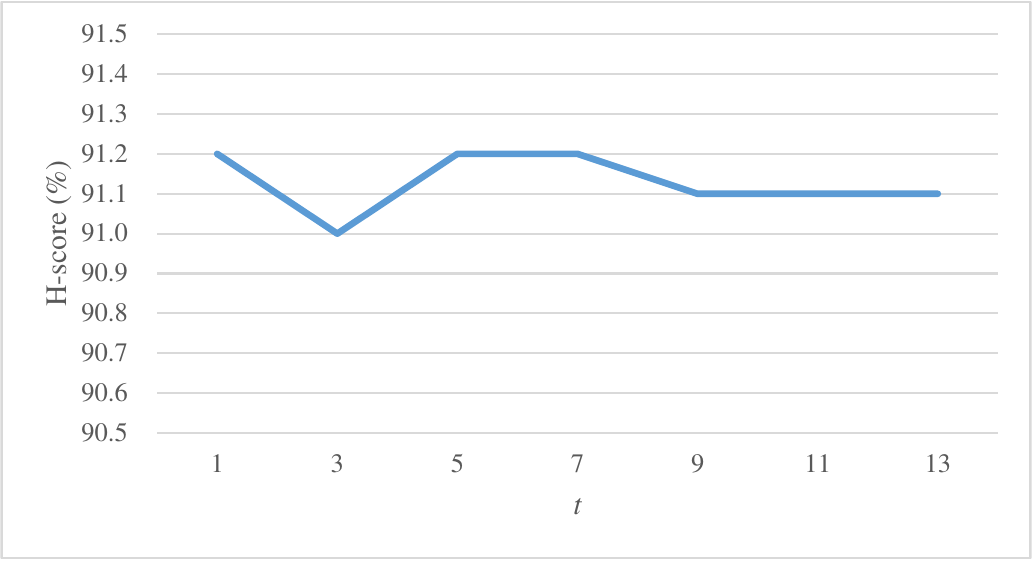}
\caption{Sensitivity of PReS to the refinement starting epoch $t$ on Office-31 under OPDA.}
\label{t_sen}
\end{figure}

We further evaluate the refinement starting epoch $t$, which determines when target-domain knowledge begins to participate in Progressive Supervision Refinement. With 15 training epochs, we vary $t$ over $\{1,3,5,7,9,11,13\}$. As shown in Fig.~\ref{t_sen}, the average H-score varies only from 91.0\% to 91.2\%, demonstrating low sensitivity to the refinement starting point. We set $t=5$ by default.

\subsection{Computational Complexity}
We analyze the additional computational complexity of PReS beyond standard target-model optimization. Before adaptation, Stable Diffusion generates one semantic-preserving counterpart for each target sample, while the black-box source model and CLIP provide predictions for supervision construction. Diffusion generation and semantic consistency verification are performed only once, with the generated samples and consistency evidence cached for subsequent adaptation. Computing the normalized Gini reliability requires $O(N_t C\log C)$ due to probability sorting, while prediction fusion and other prediction-level operations require $O(N_t C)$.

During adaptation, the dominant computation remains the forward and backward propagation of the target model. The prediction-level objectives and EMA-based pseudo-label refinement incur $O(N_t C)$ overhead, while Mixup consistency requires additional forward passes on mixed samples. After epoch $t$, Sample Partition Refinement reuses the cached semantic-consistency evidence and updates only Cross-model Consensus Verification using target predictions, without rerunning Stable Diffusion or repeating black-box queries. Overall, excluding one-time auxiliary-model inference and standard target-model optimization, PReS introduces linear overhead in $N_t$, except for the $O(N_t C\log C)$ Gini reliability computation.

\section{Conclusion}
In this paper, we proposed Progressive Reliable Supervision (PReS) for B$^2$-UniDA to address the lack of reliable target supervision under restricted source-side access and unknown label-space relationships. PReS coordinates complementary knowledge from the black-box source model, foundation models, and the evolving target model to progressively construct and refine reliable target supervision. Specifically, reliable sample- and label-level supervision is first established from complementary external knowledge and exploited through asymmetric target learning. The emerging target-domain knowledge is then fed back to refine both pseudo labels and the known/unknown sample partition, forming a closed-loop adaptation process. Extensive experiments across multiple benchmarks and diverse label-space relationships demonstrate the effectiveness of PReS under the restrictive B$^2$-UniDA setting.

\bibliographystyle{IEEEtran}
\bibliography{ref}

@ARTICLE{PHPLUDA2026Tip,
  author={Wang, Yangtao and Huang, Mingxin and Deng, Xingwei and Xie, Yanzhao and Li, Xiaocui},
  journal={IEEE Transactions on Image Processing}, 
  title={Progressive Hybrid Pseudo-Labeling for Unsupervised Domain Adaptation With Ascending Low-Rank Adaptation}, 
  year={2026},
  volume={35},
  number={},
  pages={7332-7346}}

@ARTICLE{CuiUDA2025Tip,
  author={Cui, Chaoran and Liu, Ziyi and Gong, Shuai and Zhu, Lei and Zhang, Chunyun and Liu, Hui},
  journal={IEEE Transactions on Image Processing}, 
  title={When Adversarial Training Meets Prompt Tuning: Adversarial Dual Prompt Tuning for Unsupervised Domain Adaptation}, 
  year={2025},
  volume={34},
  number={},
  pages={1427-1440}
  }

@ARTICLE{RobustSFDA2026Tip,
  author={Xiao, Yao and Wei, Pengxu and Wang, Guangrun and Liu, Cong and Lin, Liang},
  journal={IEEE Transactions on Image Processing}, 
  title={Robust Source-Free Domain Adaptation From Non-Robust Source Models}, 
  year={2026},
  volume={35},
  number={},
  pages={2350-2363}}

@ARTICLE{HGSFDA2025TIP,   
author={Jiang, Jinkun and Lv, Qingxuan and Li, Yuezun and Du, Yong and Dong, Junyu and Chen, Sheng and Yu, Hui},   
journal={IEEE Transactions on Image Processing},    title={HG-SFDA: HyperGraph Learning Meets Source-Free Unsupervised Domain Adaptation},    
year={2025},   
volume={34},   
number={},   
pages={7542-7557}
}

@ARTICLE{WSPDA2024TIP,
  author={Lan, Mengcheng and Meng, Min and Yu, Jun and Wu, Jigang},
  journal={IEEE Transactions on Image Processing}, 
  title={Learning to Discover Knowledge: A Weakly-Supervised Partial Domain Adaptation Approach}, 
  year={2024},
  volume={33},
  number={},
  pages={4090-4103}
}

@article{bommasani2021opportunities,
  title={On the opportunities and risks of foundation models},
  author={Bommasani, Rishi and Hudson, Drew A and Adeli, Ehsan and Altman, Russ and Arora, Simran and von Arx, Sydney and Bernstein, Michael S and Bohg, Jeannette and Bosselut, Antoine and Brunskill, Emma and others},
  journal={arXiv preprint arXiv:2108.07258},
  year={2021}
}

@ARTICLE{jing2023WDAN,
  author={Li, Jing and Yang, Liu and Wang, Qilong and Hu, Qinghua},
  journal={IEEE Transactions on Circuits and Systems for Video Technology}, 
  title={WDAN: A Weighted Discriminative Adversarial Network With Dual Classifiers for Fine-Grained Open-Set Domain Adaptation}, 
  year={2023},
  volume={33},
  number={9},
  pages={5133-5147},
  doi={10.1109/TCSVT.2023.3249200}
}

@inproceedings{qu2023upcycling,
  title={Upcycling models under domain and category shift},
  author={Qu, Sanqing and Zou, Tianpei and R{\"o}hrbein, Florian and Lu, Cewu and Chen, Guang and Tao, Dacheng and Jiang, Changjun},
  booktitle={Proceedings of the IEEE/CVF conference on computer vision and pattern recognition},
  pages={20019--20028},
  year={2023}
}

@article{stark2024gini,
  title   = {A Note on Sen's Representation of the Gini Coefficient:
             Revision and Repercussions},
  author  = {Stark, Oded},
  journal = {The Journal of Economic Inequality},
  volume  = {22},
  pages   = {1061--1067},
  year    = {2024},
  doi     = {10.1007/s10888-024-09623-y}
}

@article{eliazar2024beautiful,
  title   = {Beautiful Gini},
  author  = {Eliazar, Iddo},
  journal = {METRON},
  volume  = {82},
  pages   = {293--313},
  year    = {2024},
  doi     = {10.1007/s40300-024-00271-w}
}

@article{zhang2022transfer,
  title={Transfer adaptation learning: A decade survey},
  author={Zhang, Lei and Gao, Xinbo},
  journal={IEEE Transactions on Neural Networks and Learning Systems},
  volume={35},
  number={1},
  pages={23--44},
  year={2022},
  publisher={IEEE}
}

@article{li2025domain,
  title={Domain-Invariant Feature Enhancement Domain Adaptation for Cross-Scene Road Damage Detection},
  author={Li, Jie and Qu, Zhong and Yin, Xuehui},
  journal={IEEE Transactions on Circuits and Systems for Video Technology},
  year={2025},
  publisher={IEEE}
}

@article{ben2010theory,
  title={A theory of learning from different domains},
  author={Ben-David, Shai and Blitzer, John and Crammer, Koby and Kulesza, Alex and Pereira, Fernando and Vaughan, Jennifer Wortman},
  journal={Machine learning},
  volume={79},
  number={1},
  pages={151--175},
  year={2010},
  publisher={Springer}
}

@inproceedings{ganin2015unsupervised,
  title={Unsupervised domain adaptation by backpropagation},
  author={Ganin, Yaroslav and Lempitsky, Victor},
  booktitle={International conference on machine learning},
  pages={1180--1189},
  year={2015},
  organization={PMLR}
}

@inproceedings{zhang2018importance,
  title={Importance weighted adversarial nets for partial domain adaptation},
  author={Zhang, Jing and Ding, Zewei and Li, Wanqing and Ogunbona, Philip},
  booktitle={Proceedings of the IEEE conference on computer vision and pattern recognition},
  pages={8156--8164},
  year={2018}
}

@inproceedings{saito2018open,
  title={Open set domain adaptation by backpropagation},
  author={Saito, Kuniaki and Yamamoto, Shohei and Ushiku, Yoshitaka and Harada, Tatsuya},
  booktitle={Proceedings of the European conference on computer vision (ECCV)},
  pages={153--168},
  year={2018}
}

@inproceedings{panareda2017open,
  title={Open set domain adaptation},
  author={Panareda Busto, Pau and Gall, Juergen},
  booktitle={Proceedings of the IEEE international conference on computer vision},
  pages={754--763},
  year={2017}
}

@inproceedings{you2019universal,
  title={Universal domain adaptation},
  author={You, Kaichao and Long, Mingsheng and Cao, Zhangjie and Wang, Jianmin and Jordan, Michael I},
  booktitle={Proceedings of the IEEE/CVF conference on computer vision and pattern recognition},
  pages={2720--2729},
  year={2019}
}

@inproceedings{fu2020learning,
  title={Learning to detect open classes for universal domain adaptation},
  author={Fu, Bo and Cao, Zhangjie and Long, Mingsheng and Wang, Jianmin},
  booktitle={European conference on computer vision},
  pages={567--583},
  year={2020},
  organization={Springer}
}

@inproceedings{saito2021ovanet,
  title={Ovanet: One-vs-all network for universal domain adaptation},
  author={Saito, Kuniaki and Saenko, Kate},
  booktitle={Proceedings of the ieee/cvf international conference on computer vision},
  pages={9000--9009},
  year={2021}
}

@article{saito2020universal,
  title={Universal domain adaptation through self supervision},
  author={Saito, Kuniaki and Kim, Donghyun and Sclaroff, Stan and Saenko, Kate},
  journal={Advances in neural information processing systems},
  volume={33},
  pages={16282--16292},
  year={2020}
}

@inproceedings{kundu2020universal,
  title={Universal source-free domain adaptation},
  author={Kundu, Jogendra Nath and Venkat, Naveen and Babu, R Venkatesh and others},
  booktitle={Proceedings of the IEEE/CVF conference on computer vision and pattern recognition},
  pages={4544--4553},
  year={2020}
}

@article{liang2021umad,
  title={Umad: Universal model adaptation under domain and category shift},
  author={Liang, Jian and Hu, Dapeng and Feng, Jiashi and He, Ran},
  journal={arXiv preprint arXiv:2112.08553},
  year={2021}
}

@article{deng2021universal,
  title={On universal black-box domain adaptation},
  author={Deng, Bin and Zhang, Yabin and Tang, Hui and Ding, Changxing and Jia, Kui},
  journal={arXiv preprint arXiv:2104.04665},
  year={2021}
}

@inproceedings{he2025gradient,
  title={Gradient-based sample selection for black-box universal domain adaptation},
  author={He, Qiuyan and Deng, Minghua},
  booktitle={Proceedings of the AAAI Conference on Artificial Intelligence},
  volume={39},
  number={16},
  pages={17114--17122},
  year={2025}
}

@inproceedings{lai2025adu,
  title={ADU: Adaptive Detection of Unknown Categories in Black-Box Domain Adaptation},
  author={Lai, Yushan and Li, Guowen and Liang, Haoyuan and Zheng, Juepeng and Ye, Zhiyu},
  booktitle={Proceedings of the IEEE/CVF conference on computer vision and pattern recognition},
  pages={30588--30598},
  year={2025}
}

@article{qu2025glc++,
  title={Glc++: Source-free universal domain adaptation through global-local clustering and contrastive affinity learning},
  author={Qu, Sanqing and Zou, Tianpei and R{\"o}hrbein, Florian and Lu, Cewu and Chen, Guang and Tao, Dacheng and Jiang, Changjun},
  journal={IEEE Transactions on Pattern Analysis and Machine Intelligence},
  year={2025},
  publisher={IEEE}
}

@inproceedings{qu2024lead,
  title={Lead: Learning decomposition for source-free universal domain adaptation},
  author={Qu, Sanqing and Zou, Tianpei and He, Lianghua and R{\"o}hrbein, Florian and Knoll, Alois and Chen, Guang and Jiang, Changjun},
  booktitle={Proceedings of the IEEE/CVF conference on computer vision and pattern recognition},
  pages={23334--23343},
  year={2024}
}

@inproceedings{radford2021learning,
  title={Learning transferable visual models from natural language supervision},
  author={Radford, Alec and Kim, Jong Wook and Hallacy, Chris and Ramesh, Aditya and Goh, Gabriel and Agarwal, Sandhini and Sastry, Girish and Askell, Amanda and Mishkin, Pamela and Clark, Jack and others},
  booktitle={International conference on machine learning},
  pages={8748--8763},
  year={2021},
  organization={PmLR}
}

@inproceedings{sohl2015deep,
  title={Deep unsupervised learning using nonequilibrium thermodynamics},
  author={Sohl-Dickstein, Jascha and Weiss, Eric and Maheswaranathan, Niru and Ganguli, Surya},
  booktitle={International conference on machine learning},
  pages={2256--2265},
  year={2015},
  organization={pmlr}
}

@inproceedings{rombach2022high,
  title={High-resolution image synthesis with latent diffusion models},
  author={Rombach, Robin and Blattmann, Andreas and Lorenz, Dominik and Esser, Patrick and Ommer, Bj{\"o}rn},
  booktitle={Proceedings of the IEEE/CVF conference on computer vision and pattern recognition},
  pages={10684--10695},
  year={2022}
}

@article{zhou2025cross,
  title={Cross-domain Class Context Optimization for Universal Domain Adaptation},
  author={Zhou, Bo and Liu, Long and Fan, Chenyue and Zhao, Zhipeng},
  journal={IEEE Transactions on Circuits and Systems for Video Technology},
  year={2025},
  publisher={IEEE}
}

@article{wang2025double,
  title={Double-boundary awareness of shared categories for source-free universal domain adaptation},
  author={Wang, Zhijing and Guo, Ji and Sun, Xu and Luo, Yi and Chen, Aiguo},
  journal={Neurocomputing},
  pages={131473},
  year={2025},
  publisher={Elsevier}
}

@article{lu2026exploring,
  title={Exploring Generic Knowledge and Reactivating Source Model for Source-free Universal Domain Adaptation},
  author={Lu, Yuwu and Lan, Yifan and Yang, Huan and Lai, Zhihui and Li, Xuelong},
  journal={IEEE Transactions on Multimedia},
  year={2026},
  publisher={IEEE}
}

@inproceedings{he2016deep,
  title={Deep residual learning for image recognition},
  author={He, Kaiming and Zhang, Xiangyu and Ren, Shaoqing and Sun, Jian},
  booktitle={Proceedings of the IEEE conference on computer vision and pattern recognition},
  pages={770--778},
  year={2016}
}

@article{hinton2015distilling,
  title={Distilling the knowledge in a neural network},
  author={Hinton, Geoffrey and Vinyals, Oriol and Dean, Jeff},
  journal={arXiv preprint arXiv:1503.02531},
  year={2015}
}

@inproceedings{liang2020we,
  title={Do we really need to access the source data? source hypothesis transfer for unsupervised domain adaptation},
  author={Liang, Jian and Hu, Dapeng and Feng, Jiashi},
  booktitle={International conference on machine learning},
  pages={6028--6039},
  year={2020},
  organization={PMLR}
}

@article{krause2010discriminative,
  title={Discriminative clustering by regularized information maximization},
  author={Krause, Andreas and Perona, Pietro and Gomes, Ryan},
  journal={Advances in neural information processing systems},
  volume={23},
  year={2010}
}

@article{li2024hyunida,
  title={Hyunida: Breaking label set constraints for universal domain adaptation in cross-scene hyperspectral image classification},
  author={Li, Qingmei and Wen, Yibin and Zheng, Juepeng and Zhang, Yuxiang and Fu, Haohuan},
  journal={IEEE Transactions on Geoscience and Remote Sensing},
  volume={62},
  pages={1--15},
  year={2024},
  publisher={IEEE}
}

@inproceedings{long2015learning,
  title={Learning transferable features with deep adaptation networks},
  author={Long, Mingsheng and Cao, Yue and Wang, Jianmin and Jordan, Michael},
  booktitle={International conference on machine learning},
  pages={97--105},
  year={2015},
  organization={PMLR}
}

@inproceedings{cui2020gradually,
  title={Gradually vanishing bridge for adversarial domain adaptation},
  author={Cui, Shuhao and Wang, Shuhui and Zhuo, Junbao and Su, Chi and Huang, Qingming and Tian, Qi},
  booktitle={Proceedings of the IEEE/CVF conference on computer vision and pattern recognition},
  pages={12455--12464},
  year={2020}
}

@inproceedings{cao2018partial,
  title={Partial transfer learning with selective adversarial networks},
  author={Cao, Zhangjie and Long, Mingsheng and Wang, Jianmin and Jordan, Michael I},
  booktitle={Proceedings of the IEEE conference on computer vision and pattern recognition},
  pages={2724--2732},
  year={2018}
}

@inproceedings{cao2019learning,
  title={Learning to transfer examples for partial domain adaptation},
  author={Cao, Zhangjie and You, Kaichao and Long, Mingsheng and Wang, Jianmin and Yang, Qiang},
  booktitle={Proceedings of the IEEE/CVF conference on computer vision and pattern recognition},
  pages={2985--2994},
  year={2019}
}

@inproceedings{saenko2010adapting,
  title={Adapting visual category models to new domains},
  author={Saenko, Kate and Kulis, Brian and Fritz, Mario and Darrell, Trevor},
  booktitle={European conference on computer vision},
  pages={213--226},
  year={2010},
  organization={Springer}
}

@inproceedings{venkateswara2017deep,
  title={Deep hashing network for unsupervised domain adaptation},
  author={Venkateswara, Hemanth and Eusebio, Jose and Chakraborty, Shayok and Panchanathan, Sethuraman},
  booktitle={Proceedings of the IEEE conference on computer vision and pattern recognition},
  pages={5018--5027},
  year={2017}
}

@article{peng2017visda,
  title={Visda: The visual domain adaptation challenge},
  author={Peng, Xingchao and Usman, Ben and Kaushik, Neela and Hoffman, Judy and Wang, Dequan and Saenko, Kate},
  journal={arXiv preprint arXiv:1710.06924},
  year={2017}
}

@inproceedings{peng2019moment,
  title={Moment matching for multi-source domain adaptation},
  author={Peng, Xingchao and Bai, Qinxun and Xia, Xide and Huang, Zijun and Saenko, Kate and Wang, Bo},
  booktitle={Proceedings of the IEEE/CVF international conference on computer vision},
  pages={1406--1415},
  year={2019}
}

@inproceedings{li2021domain,
  title={Domain consensus clustering for universal domain adaptation},
  author={Li, Guangrui and Kang, Guoliang and Zhu, Yi and Wei, Yunchao and Yang, Yi},
  booktitle={Proceedings of the IEEE/CVF conference on computer vision and pattern recognition},
  pages={9757--9766},
  year={2021}
}

@inproceedings{li2025esbn,
  title={ESBN: Estimation Shift of Batch Normalization for Source-free Universal Domain Adaptation},
  author={Li, Jiao and Su, Houcheng and Wang, Bingli and Min, Yuandong and Wang, Mengzhu and Yin, Nan and Wang, Shanshan and Guo, Jingcai},
  booktitle={Proceedings of the Thirty-Fourth International Joint Conference on Artificial Intelligence},
  pages={1386--1394},
  year={2025}
}

@article{paszke2017automatic,
  title={Automatic differentiation in pytorch},
  author={Paszke, Adam and Gross, Sam and Chintala, Soumith and Chanan, Gregory and Yang, Edward and DeVito, Zachary and Lin, Zeming and Desmaison, Alban and Antiga, Luca and Lerer, Adam},
  year={2017}
}

@inproceedings{deng2009imagenet,
  title={Imagenet: A large-scale hierarchical image database},
  author={Deng, Jia and Dong, Wei and Socher, Richard and Li, Li-Jia and Li, Kai and Fei-Fei, Li},
  booktitle={2009 IEEE conference on computer vision and pattern recognition},
  pages={248--255},
  year={2009},
  organization={Ieee}
}

@inproceedings{zeng2025kabon,
  title={KABON: Knowledge Aggregation with Vision-Language Model for Black-Box Open-Set Domain Adaptation},
  author={Zeng, Zhixin and Zhang, Yusen and Wang, Ji},
  booktitle={ICASSP 2025-2025 IEEE International Conference on Acoustics, Speech and Signal Processing (ICASSP)},
  pages={1--5},
  year={2025},
  organization={IEEE}
}

@article{van2008visualizing,
  title={Visualizing data using t-SNE.},
  author={Van der Maaten, Laurens and Hinton, Geoffrey},
  journal={Journal of machine learning research},
  volume={9},
  number={11},
  year={2008}
}

@inproceedings{selvaraju2017grad,
  title={Grad-cam: Visual explanations from deep networks via gradient-based localization},
  author={Selvaraju, Ramprasaath R and Cogswell, Michael and Das, Abhishek and Vedantam, Ramakrishna and Parikh, Devi and Batra, Dhruv},
  booktitle={Proceedings of the IEEE international conference on computer vision},
  pages={618--626},
  year={2017}
}

@inproceedings{li2022blip,
  title={Blip: Bootstrapping language-image pre-training for unified vision-language understanding and generation},
  author={Li, Junnan and Li, Dongxu and Xiong, Caiming and Hoi, Steven},
  booktitle={International conference on machine learning},
  pages={12888--12900},
  year={2022},
  organization={PMLR}
}

@article{alayrac2022flamingo,
  title={Flamingo: a visual language model for few-shot learning},
  author={Alayrac, Jean-Baptiste and Donahue, Jeff and Luc, Pauline and Miech, Antoine and Barr, Iain and Hasson, Yana and Lenc, Karel and Mensch, Arthur and Millican, Katherine and Reynolds, Malcolm and others},
  journal={Advances in neural information processing systems},
  volume={35},
  pages={23716--23736},
  year={2022}
}

@inproceedings{singh2022flava,
  title={Flava: A foundational language and vision alignment model},
  author={Singh, Amanpreet and Hu, Ronghang and Goswami, Vedanuj and Couairon, Guillaume and Galuba, Wojciech and Rohrbach, Marcus and Kiela, Douwe},
  booktitle={Proceedings of the IEEE/CVF conference on computer vision and pattern recognition},
  pages={15638--15650},
  year={2022}
}

\vspace{0 mm}

\end{document}